%% file: iclr2027_conference.tex
\documentclass{article} 
\usepackage{iclr2027_conference,times}

\input{math_commands.tex}

\usepackage{hyperref}
\usepackage{url}

\usepackage{algorithm}
\usepackage{algorithmic}
\usepackage{amsmath}
\usepackage{amssymb}

\usepackage{multirow}
\usepackage[table]{xcolor}

\usepackage{graphicx}
\usepackage{booktabs}
\usepackage{wrapfig}
\usepackage{float}
\usepackage{colortbl}
\usepackage{caption}
\usepackage{tikz}
\usepackage{makecell}

\usepackage{marvosym}
\usepackage{fontawesome5}

\newcommand{\INPUT}{\item[\textbf{Input:}]}
\newcommand{\OUTPUT}{\item[\textbf{Output:}]}

\definecolor{best}{HTML}{FFCC99}
\definecolor{best2}{HTML}{BBDEFB}

\title{FoundDSR: A Generalizable Foundation \\ Model with Guided 2D Gaussian Splatting \\ for Depth Super-Resolution}

\author{Zhengxue Wang$^{1}$~ Zhiqiang Yan$^{2}$\textsuperscript{\Letter}~ Yuan Wu$^{1}$~ Guangwei Gao$^{1}$~ Xiang Li$^{3}$~ Jian Yang$^{1,4}$\textsuperscript{\Letter} \\
\small{$^1$PCA Lab, Nanjing University of Science and Technology}\hspace{1em}
\\
\small{$^2$National University of Singapore}\hspace{1em}
\small{$^3$Nankai University}\hspace{1em}
\small{$^4$Nanjing University}
\\
{\tt \small{zxwang@njust.edu.cn, yanzq@nus.edu.sg}}
}

\iclrfinalcopy 

\begin{document}

\maketitle

\footnotetext[1]{\Letter~Corresponding authors}

\begin{abstract}
We introduce FoundDSR, a generalizable foundation model for robust depth reconstruction across unseen data distributions using RGB-D pairs.
FoundDSR begins with a guided 2D Gaussian Splatting strategy to model depth representations with Gaussian primitives. This strategy employs high-resolution RGB as prompts to optimize the Gaussian parameters, thereby encouraging each Gaussian primitive to anisotropically deform along high-frequency structural directions. 
The resulting Gaussian-upsampled representations are then mapped to high-resolution depth through an effective depth reconstruction branch. 
Furthermore, to mitigate training instability and bias toward dominant sources caused by distribution gaps in large-scale heterogeneous data, we introduce heterogeneous federated learning that allocates each data source to an independent client for local optimization and global aggregation. This design effectively endows FoundDSR with stable scalability to diverse and large-scale training data. 
Extensive zero-shot evaluations on synthetic, real-world, arbitrary-scale, and noisy conditions demonstrate that FoundDSR consistently outperforms existing state-of-the-art approaches, confirming its strong robustness and generalization to unknown scenes. Codes are available at \url{https://github.com/yanzq95/FoundDSR}
\end{abstract}

\section{Introduction}
\label{sec:intro}

Depth maps capture rich geometric information, making them essential for various computer vision tasks, including augmented reality~\cite{zhao2023spherical, wang2023rgb}, virtual reality~\cite{ye2025semantics}, 3D reconstruction~\cite{yan2026event, sun2021learning}, and robotic navigation~\cite{yuan2023structure, song2020channel}. Due to the limitations of sensor technology and imaging environments, consumer-grade devices typically acquire low-resolution (LR) depth corrupted by unknown degradations. To this end, numerous depth super-resolution (DSR) methods~\cite{yang2022codon, chen2024intrinsic, zheng2025decoupling} have been proposed to restore accurate high-resolution (HR) depth from their LR inputs, with RGB employed as guidance to facilitate depth recovery.

However, real-world scenarios are inherently variable, with acquisition environments and target objects often diverging significantly from training data. Moreover, different types of depth sensors often exhibit distinct noise distributions and degradation patterns during imaging. Most existing DSR methods~\cite{wang2023learning, wang2024sgnet} rely on small-scale and homogeneous data, resulting in suboptimal zero-shot generalization to domains outside the training distribution.
Although recent works~\cite{yan2025ducos, wang2026scene, wang2026graph} leverage geometric priors from pretrained vision foundation models to improve robustness, they are still fine-tuned on limited and less diverse data, failing to overcome the zero‑shot generalization bottleneck in real‑world scenes.

In this paper, we propose FoundDSR, a novel foundation model for DSR trained on million-scale data to significantly improve zero-shot generalization across unseen domains. 
To enable arbitrary-scale depth restoration, we develop a guided 2D Gaussian Splatting strategy that models depth features with continuous Gaussian primitives. Specifically, this strategy employs RGB gradients as 
\begin{wrapfigure}{r}{0.50\textwidth}
\centering
\includegraphics[width=\linewidth]{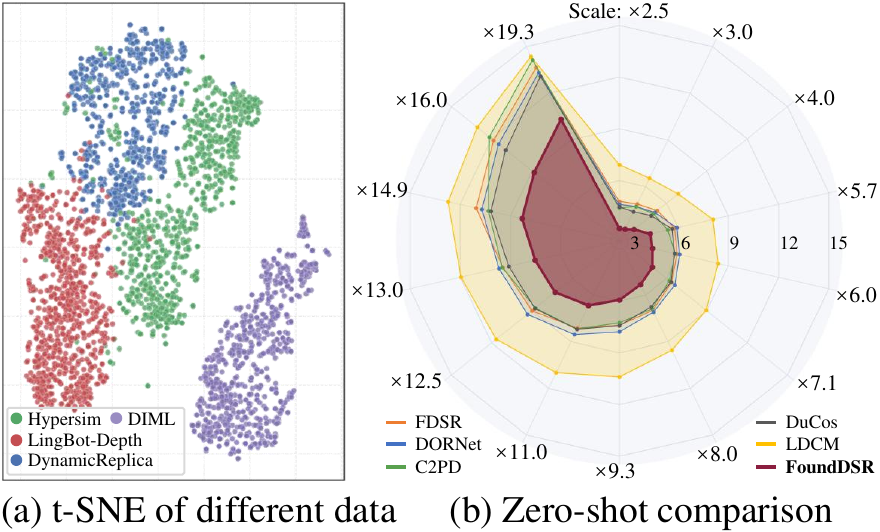}
\setlength{\abovecaptionskip}{-4pt}
\caption{(a) t-SNE visualization of feature embeddings from different data sources. (b) Average RMSE across all benchmarks at different scales.}
\label{fig:dist_RMSEComp}
\vspace{-12pt}
\end{wrapfigure}
prompts to adaptively modulate both the covariance and spatial position of each Gaussian primitive, driving them to anisotropically deform in orientation, shape, and position along high-frequency structural directions. This anisotropic design facilitates accurate depth reconstruction at arbitrary upsampling factors. Subsequently, we introduce a simple yet effective depth reconstruction branch. Eschewing complex module designs, it adopts a straightforward encoder-decoder architecture to restore accurate HR depth from the Gaussian-upsampled depth representations, while maintaining low inference latency.

Furthermore, Fig.~\ref{fig:dist_RMSEComp}(a) visualizes the embedding distributions across different data sources, revealing pronounced distribution gaps in large-scale heterogeneous data. During mixed-data training, such discrepancies can bias optimization toward dominant sources with larger proportions and destabilize training. To address this issue, we introduce a heterogeneous federated learning approach that assigns each data source to an independent client for local optimization. Subsequently, the server periodically aggregates and redistributes the client weights. This mechanism effectively mitigates cross-source optimization interference, significantly improving both scalability and training stability on large-scale heterogeneous datasets.

Fig.~\ref{fig:dist_RMSEComp}(b) compares the generalization performance of our FoundDSR with previous depth foundation models and DSR approaches across multiple scales. These results demonstrate that our method exhibits superior zero-shot generalization to unseen scenarios, achieving an average RMSE reduction of $20.66\%$ over the second-best method across all scales.

In summary, our contributions are as follows:
\begin{itemize}
    \item For the first time, we propose a novel DSR foundation model that is trained on large-scale heterogeneous data, substantially improving zero-shot generalization to unseen scenes.
    \item We present a guided 2D Gaussian Splatting that leverages high-resolution RGB as prompts to anisotropically deform Gaussian primitives along high-frequency structural directions, effectively modeling depth representations at arbitrary scales.
    \item We introduce heterogeneous federated learning to mitigate training instability and bias toward dominant data sources induced by distribution gaps across large-scale multi-source data, enabling stable scaling to larger and more diverse training data.  
    \item Extensive experiments across various datasets demonstrate the superior zero-shot generalization and robustness of our FoundDSR over existing state-of-the-art approaches.
\end{itemize}

\section{Related Work}
\label{sec:rw}

\subsection{Depth Super-Resolution}
Recent deep learning-based DSR methods~\cite{zhong2025dual, zhong2026dual, wang2026spatiotemporal} mainly focus on RGB-D feature fusion, exploring strategies such as shared and modality-specific feature decomposition~\cite{deng2020deep}, frequency decoupling~\cite{he2021towards}, and guided filtering~\cite{zhong2023deep, wang2026scene}.  For example, DCTNet~\cite{zhao2022discrete} disentangles shared and private RGB-D features in Euclidean and spherical feature spaces to suppress depth-irrelevant cues during multi-modal feature fusion. SGNet~\cite{wang2024sgnet} proposes a structure‑guided DSR network that simultaneously extracts and aggregates the high‑ and low‑frequency information from multi-modal inputs across the spatial, gradient, and frequency domains. DKN~\cite{kim2021deformable} incorporates a deformable mechanism into joint image filtering, adaptively transferring RGB features to depth features. More recently, several methods~\cite{wang2025dornet, yan2025ducos} have explored improving the generalization of DSR models. For instance, DuCos~\cite{yan2025ducos} presents a dual-constraint network that leverages a foundation model for depth estimation as a prompt to enhance generalization across different scenes. However, these methods still rely on fine-tuning on small-scale datasets, which limits their performance to unseen domains. To address this issue, we introduce a novel DSR foundation model trained on large-scale heterogeneous data, which demonstrates superior robustness and zero-shot generalization.

\subsection{Vision Foundation Model}
Foundation models have achieved groundbreaking progress in many computer vision fields, such as image segmentation and depth estimation. Their core principle is to improve generalization by scaling up training data rather than introducing increasingly intricate module designs. For instance, SAM~\cite{kirillov2023segment} constructs a large-scale segmentation dataset and introduces the first segment anything model, achieving remarkable zero-shot segmentation performance. Recently, Depth Anything~\cite{yang2024depth} introduces the first foundation model for monocular depth estimation, jointly trained on large-scale labeled and unlabeled data. This has inspired a subsequent line of foundation models~\cite{ke2024repurposing, hu2024metric3d} for depth estimation. For instance, PromptDA~\cite{lin2025prompting} employs low-cost LiDAR as a prompt for the decoder of a depth estimation foundation model, improving metric depth prediction. Unlike these foundation models, we introduce a novel DSR foundation model that leverages federated learning and guided 2D Gaussian Splatting, significantly improving zero-shot generalization across diverse scenes and arbitrary upsampling factors.

\subsection{Gaussian Splatting}
Gaussian Splatting (GS) has shown strong potential for image representation. 3D GS~\cite{kerbl20233d, xu2025depthsplat} combines explicit Gaussian representations with differentiable rasterization, enabling efficient 3D reconstruction and novel view synthesis. However, 3D GS typically depends on accurate camera intrinsics and extrinsics for projection and rendering, restricting its use when calibration data is unavailable. Recently, 2D GS~\cite{zeng2025instant, chen2025generalized}  has been introduced, enabling direct modeling of geometric structures within the pixel coordinate system without requiring explicit 3D projection transformations. For example, GaussianImage~\cite{zhang2024gaussianimage} proposes a pioneering 2D GS paradigm for image representation and compression, utilizing 2D Gaussians instead of 3D ones to substantially reduce the storage requirements of Gaussian representation. Different from these single-modal approaches, we introduce a cross‑modal guided 2D GS. This strategy leverages the first‑order gradients of high‑resolution RGB as structural prompts to optimize depth Gaussian primitives, thereby enhancing the accuracy of depth geometry.

\section{Method}
\label{sec:method}

\subsection{Preliminaries: 2D Gaussian Splatting}

Given an image $\boldsymbol Z$, we model it as a continuous spatial representation using a set of 2D Gaussians~\cite{peng2026pixel}, where $\boldsymbol{Z}(x,y)$ at any continuous position $(x,y)$ is obtained:
\begin{equation}\label{eq:ori_gs2}
   \boldsymbol Z(x,y)=\sum_{m=1}^{M} \mathcal{G}_{m} (x, y),\quad  \mathcal{G}_{m} (x,y)=\boldsymbol c_{rgb,m} \cdot \frac{1}{2\pi |\boldsymbol \Sigma_{m}| } \exp\left ( -\frac{1}{2} \boldsymbol d^{\top }_{m} \boldsymbol \Sigma^{-1}_{m} \boldsymbol d_{m} \right ) ,
\end{equation}
where $M$ represents the number of Gaussians. Each Gaussian kernel $\mathcal{G}_{m}$ is parameterized by its spatial position $\boldsymbol \mu_{m} =(\mu _{x,m}, \mu _{y,m})\top $, covariance matrix $\boldsymbol \Sigma_{m} $, and color coefficients $\boldsymbol c_{rgb,m}\in [0,1]^{3} $. $\boldsymbol d_{m}=(x-\mu _{x,m}, y-\mu _{y,m})\top$ denotes the displacement vector from the Gaussian center to the query point $(x, y)$. The covariance matrix $\boldsymbol \Sigma_{m} $ can be further factorized into:
\begin{equation}\label{eq:ori_sigma}
\boldsymbol \Sigma_{m}=\boldsymbol R_{m}\!\begin{bmatrix}
  \sigma_{x,m}^{2}\! &0   \\
  0\! &\sigma_{y,m}^{2} 
\end{bmatrix}\! \boldsymbol R^{\top }_{m}, \quad \boldsymbol R_{m}=\begin{bmatrix}
  \cos (\theta_{m})\! &-\sin(\theta_{m}) \\
  \sin(\theta_{m})\! &\cos (\theta_{m})
\end{bmatrix},
\end{equation}
where $\sigma_{x,m}$ and $\sigma_{y,m}$ are the standard deviations that control the shape of each Gaussian primitive. $\boldsymbol R_{m}$ is the rotation matrix and $\theta_{m}$ is the rotation angle, which together determine the orientation.

\begin{figure}[t]
\centering
\includegraphics[width=0.91\columnwidth]{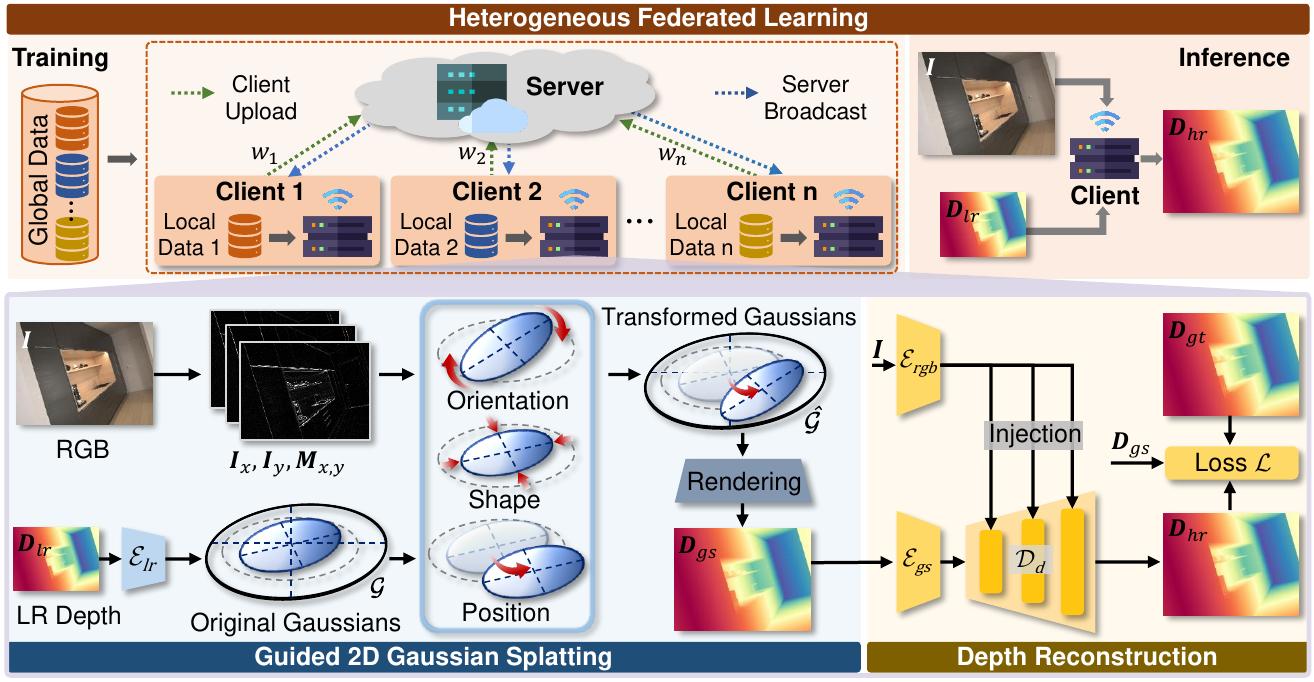}\\
\vspace{-4pt}
\caption{Pipeline of FoundDSR. In federated learning, global data are distributed to independent clients for local optimization, while the server aggregates and redistributes model weights. For each client, LR depth is encoded into initial Gaussian $\mathcal{G}$, which are modulated by RGB gradients for anisotropic structural deformation. The transformed Gaussian $\mathcal{\hat{G} }$ are then rendered into $\boldsymbol D_{gs}$. Finally, $\boldsymbol D_{gs}$ and RGB $\boldsymbol I$ are fed into a depth reconstruction to predict HR depth $\boldsymbol D_{HR}$.}\label{fig:pip}
\vspace{-6pt}
\end{figure}

\subsection{Overall Framework}
As shown in Fig.~\ref{fig:pip}, FoundDSR incorporates heterogeneous federated learning to collaboratively optimize large-scale RGB-D data, mitigating the interference caused by heterogeneous data. All clients share the same architecture, each comprising a guided 2D GS branch and a depth reconstruction branch. Given an LR depth $\boldsymbol D_{lr}\in R^{h\times w\times 1} $, where $h$ and $w$ denote its  height and width, the guided 2D GS first employs an encoder $\mathcal{E} _{lr} $ to predict initial Gaussians $\mathcal{G}$. Then, RGB gradients $\boldsymbol I_{x}$, $\boldsymbol I_{y}$, and $\boldsymbol M _{x,y}$ are utilized to modulate the covariance and position of $\mathcal{G}$,  adaptively deforming each Gaussian in orientation, shape, and position along high-frequency structural directions, yielding the transformed Gaussians $\mathcal{\hat{G} }$. Next, $\mathcal{\hat{G} }$ is rendered into an upsampled depth representation $\boldsymbol D_{gs}$. Finally, a depth reconstruction branch with dual encoders and a single decoder fuses the RGB $\boldsymbol I\in R^{sh\times sw\times 3} $ and $\boldsymbol D_{gs}$ to predict the HR depth $\boldsymbol D_{hr}\in R^{sh\times sw\times 1} $, where $s$ is the upsampling factor.

\subsection{Guided 2D Gaussian Splatting}
As illustrated in the blue region of Fig.~\ref{fig:pip}, our guided 2D GS leverages RGB structures as prompts for Gaussian prediction and modulation.   For clarity, we omit the Gaussian index $m$ in the following formulations. Specifically, the LR depth $\boldsymbol D_{lr}$ is first fed into the network $\mathcal{E}_{lr}$ to predict the original Gaussian primitives $\mathcal{G}$, as defined in Eq.~\ref{eq:ori_gs2}, where $\mathcal{E}_{lr}$ consists of multiple MLP heads that separately estimate different Gaussian parameters. Then, we compute the RGB gradients along the $x$ and $y$ directions (denoted as $\boldsymbol I_{x}$ and $\boldsymbol I_{y}$) and their corresponding magnitudes $\boldsymbol M_{x,y}$:
\begin{equation}\label{eq:gradient}
   \boldsymbol I_{x}=\frac{\partial \boldsymbol I}{\partial x} , \quad\boldsymbol I_{y}=\frac{\partial \boldsymbol I}{\partial y}, \quad \boldsymbol M_{x,y}=\sqrt{\boldsymbol I_{x}^{2} + \boldsymbol I_{y}^{2}} .
\end{equation}
Next, $\boldsymbol I_{x}$, $\boldsymbol I_{y}$, and $\boldsymbol M_{x,y}$ are jointly employed to modulate the position $\boldsymbol \mu$, rotation angle $\theta$, and covariance matrix $\boldsymbol \Sigma$ of $\mathcal{G}$.  This strategy enables the orientations, shapes, and positions of the 2D Gaussians to undergo adaptive anisotropic deformation along the gradient directions, facilitating the recovery of high-frequency geometric information in rendered depth. The modulated position is:
\begin{equation}\label{eq:mu_hat}
   \hat{\boldsymbol \mu } =\begin{bmatrix}
 \hat{\mu }_{x}\\
 \hat{\mu }_{y}
\end{bmatrix},\left\{\begin{matrix}
 \hat{\mu }_{x}=\mu_{x}+\frac{\boldsymbol I_{x}}{\boldsymbol M_{x,y}+\epsilon }  \\
 \hat{\mu }_{y}=\mu_{y}+\frac{\boldsymbol I_{y}}{\boldsymbol M_{x,y}+\epsilon } 
\end{matrix}\right.,
\end{equation}
where $\epsilon$ is a small constant for numerical stability. The modulated angle $\hat{\theta } $ and matrix $\hat{\boldsymbol R} $ are:
\begin{equation}\label{eq:rotation}
\hat{\boldsymbol R}  = \begin{bmatrix}
  \cos (\hat{\theta} )   &-\sin(\hat{\theta} ) \\
  \sin(\hat{\theta} )   &\cos (\hat{\theta} )
\end{bmatrix}, \quad\begin{cases}
 \phi  = \text{arctan2}(\boldsymbol I_{y}, \boldsymbol I_{x}) + \frac{\pi }{2}  \\
 \hat{\theta }=\theta +\lambda (\phi-\theta)
\end{cases},
\end{equation}
\begin{wrapfigure}{r}{0.48\textwidth}
\centering
\includegraphics[width=\linewidth]{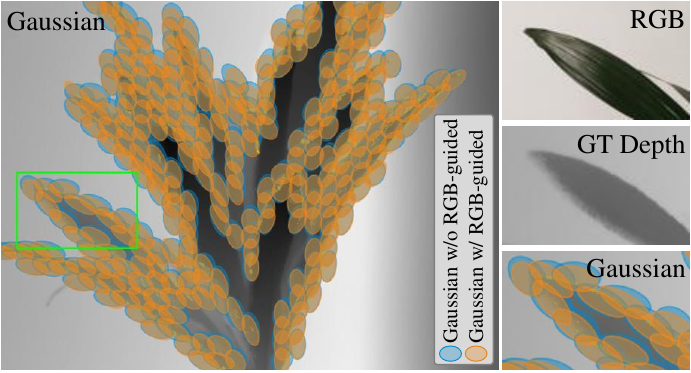}
\setlength{\abovecaptionskip}{-4pt}
\caption{Visualization of Gaussian primitive modulation with and without RGB guidance.}
\label{fig:gaussians}
\vspace{-18pt}
\end{wrapfigure}
where $\lambda$ is a learnable constant controlling the rotation toward the edge-tangent direction. Based on $\hat{\boldsymbol R}$, we obtain the covariance matrix:
\begin{equation}\label{eq:sigma_hat}
 \hat{\boldsymbol \Sigma}\! =\!\hat{\boldsymbol R}  \begin{bmatrix}
  \sigma_{x}^{2}(1+\alpha \boldsymbol M_{x,y})  &0   \\
  0  &\frac{\sigma_{y}^{2} }{1+\beta  \boldsymbol M_{x,y}} 
\end{bmatrix} \! \hat{\boldsymbol R} ^{\top }\!,
\end{equation}
where $\alpha$ and $\beta$ are learnable parameters that scale the Gaussian along tangent and normal directions.

The transformed Gaussian $\mathcal{\hat{G} }$ are formulated as:  
\begin{equation}\label{eq:gs_hat}
 \mathcal{\hat{G} }=\boldsymbol c_{d} \cdot \frac{1}{2\pi |\hat{\boldsymbol \Sigma}| } \exp\left ( -\frac{1}{2} \hat{\boldsymbol d} ^{\top } \hat{\boldsymbol \Sigma}^{-1} \hat{\boldsymbol d} \right ) ,
\end{equation}
where the displacement vector $\hat{\boldsymbol d}=(x-\hat{\mu }_{x}, y-\hat{\mu }_{y})\top$. $\boldsymbol c_{d}$ denotes the depth coefficient from the prediction head in $\mathcal{E}_{lr}$. Finally, following existing 2D Gaussian rendering methods~\cite{ye2025gsplat, peng2026pixel}, $\mathcal{\hat{G} }$ is rendered into the upsampled depth representation $\boldsymbol D_{gs}$. Fig.~\ref{fig:gaussians} visualizes the Gaussian primitives predicted with and without our RGB-guided strategy. The results show that our method effectively modulates the positions, shapes, and orientations of the Gaussian primitives, enabling them to better align with the depth structures.

\subsection{Depth Reconstruction}
\begin{wrapfigure}{r}{0.51\textwidth}
\vspace{-10pt}
\begin{minipage}{\linewidth}
\hrule
\vspace{3pt}
\refstepcounter{algorithm}
\noindent\textbf{Algorithm \thealgorithm:} Model Parameter Aggregation
\label{alg:client_weighted_agg}

\vspace{3pt}
\hrule

\begin{algorithmic}[1]
\INPUT Global model parameters $\psi^t$ at round $t$, local model parameters $\{\hat{\psi}_i^t\}_{i=1}^n$ from $n$ clients
\OUTPUT Updated global model parameters $\psi^{t+1}$

\STATE \textbf{Step 1: Compute client updates}
\FOR{each client $i = 1$ to $n$}
    \STATE $\Delta_i^t \leftarrow \hat{\psi}_i^t - \psi^t$
\ENDFOR

\STATE \textbf{Step 2: Compute average update direction}
\STATE $\bar{\Delta}^t \leftarrow \frac{1}{n} \sum_{i=1}^n \Delta_i^t$

\STATE \textbf{Step 3: Compute directional consistency and aggregation weights}
\FOR{each client $i = 1$ to $n$}
    \STATE $c_i^t \leftarrow
    \frac{\Delta_i^t \cdot \bar{\Delta}^t}
    {\|\Delta_i^t\| \|\bar{\Delta}^t\|}$
    \COMMENT{Direction consistency of client parameter updates}

    \STATE $w_i^t \leftarrow
    \frac{\exp(c_i^t / \tau)}
    {\sum_{j=1}^n \exp(c_j^t / \tau)}$
    \COMMENT{Aggregation weight with temperature coefficient $\tau$}
\ENDFOR

\STATE \textbf{Step 4: Aggregate global model}
\STATE $\psi^{t+1} \leftarrow
\sum_{i=1}^n w_i^t \, \hat{\psi}_i^t$

\RETURN $\psi^{t+1}$
\end{algorithmic}

\hrule

\end{minipage}
\vspace{-10pt}
\end{wrapfigure}
As shown in the yellow region of Fig.~\ref{fig:pip}, our depth reconstruction branch aligns with the prevailing paradigm of most depth foundation models~\cite{yang2024depth, yang2024depthv2, lin2025prompting}, adopting a simple yet effective encoder–decoder network to reconstruct accurate HR depth $\boldsymbol D_{hr}$ from the RGB guidance $\boldsymbol I$ and the $\boldsymbol D_{gs}$ predicted by the guided 2D GS. Without relying on intricate module designs, we first employ two DINOv2~\cite{oquab2023dinov2} encoders, $\mathcal{E} _{rgb} $ and $\mathcal{E} _{gs} $, to extract RGB features and depth features. Then, a DPT-based decoder~\cite{ranftl2021vision} $\mathcal{D} _{d} $ is introduced, which aggregates multi-scale RGB-D features from the encoders through a straightforward addition operation, progressively recovering spatial details and ultimately yielding $\boldsymbol D_{hr}$.

In this work, following prior foundation models~\cite{lin2025prompting, yu2026large}, we adopt the ViT-L variant of DINOv2 as the default encoders for both $\mathcal{E}_{rgb}$ and $\mathcal{E}_{gs}$ in our FoundDSR. To reduce computational cost while preserving competitive generalization, we further introduce a lightweight FoundDSR-S, which replaces ViT-L with ViT-S and keeps the remaining architecture unchanged.  

\subsection{Heterogeneous Federated Learning}

As depicted in the orange region of Fig.~\ref{fig:pip}, we introduce a heterogeneous federated learning strategy to mitigate optimization interference caused by distribution discrepancies across diverse data sources.
%
Specifically, the large-scale global training data are partitioned by source into multiple local subsets, each of which is used to independently optimize a client. All clients share the same DSR architecture, consisting of our proposed guided 2D GS and depth reconstruction network, and follow an identical training strategy to ensure consistent local optimization.  
In each communication round, the model parameters learned by individual clients are uploaded to the server for adaptive aggregation~\cite{shi2025fedawa}. The aggregated parameters are then redistributed to all clients, enabling iterative refinement and collaborative training across heterogeneous data sources.

Alg.~\ref{alg:client_weighted_agg} outlines the aggregation strategy for client model parameters. At round $t$, given the global model parameters $\psi^t$ and the locally trained parameters $\{\hat{\psi}_i^t\}_{i=1}^n$ from $n$ clients (where $n$ equals the number of data sources, set to $4$ by default), we first compute each client update, yielding $\Delta_i^t$. The average update $\bar{\Delta}^t$ is then estimated based on all $\Delta_i^t$. To improve update stability, we further evaluate the consistency between each client's update direction and $\bar{\Delta}^t$, producing consistency scores $c_i^t$. These scores are subsequently used to assign aggregation weights $w_i^t$ to each client, ensuring that clients whose update directions are better aligned with $\bar{\Delta}^t$ receive higher weights, thereby effectively improving aggregation stability. Finally, the aggregated parameters $\psi^{t+1}$ are broadcast to all clients for the next communication round.

\subsection{Loss Function}
To optimize our FoundDSR, we introduce a training loss $\mathcal{L} $ composed of a reconstruction loss $\mathcal{L}_{rec}$ and a Gaussian loss $\mathcal{L}_{gs}$. The former supervises each client to recover the final HR depth $\boldsymbol D_{hr}$, while the latter constrains the guided 2D Gaussian Splatting process to preserve structural fidelity in the rendered depth representation $\boldsymbol D_{gs}$:
\begin{equation}
    \mathcal{L} =\underbrace{{\textstyle \sum_{q\in \mathbb{Q}}}||\boldsymbol D_{gt}^{q}-\boldsymbol D_{hr}^{q}||_{1}}_{\text{Reconstruction loss $\mathcal{L}_{rec}$}}+ \rho\underbrace{ \! \left (\! {\textstyle \sum_{q\in \mathbb{Q}}}||\boldsymbol D_{gt}^{q}\!-\!\boldsymbol D_{gs}^{q}||_{1} \!+\!\!\left (\! 1\!-\!\frac{\sigma_{gs,gt}+C}{\sigma_{gs}\sigma_{gt}+C} \! \right ) \!\! \right ) }_{\text{Gaussian loss $\mathcal{L}_{gs}$}},
    \label{eq:loss}
\end{equation}
where $\boldsymbol D_{gt}$ represents the ground-truth (GT) depth, and $\mathbb{Q}$ is the valid pixel set of $\boldsymbol D_{gt}$. $||\cdot ||_{1}$ denotes the $L_{1}$ norm.  $\rho=0.1$ is a balancing coefficient. $\sigma_{gs}$ and $\sigma_{gt}$ are the standard deviations of $\boldsymbol D_{gs}$ and $\boldsymbol D_{gt}$, respectively, while $\sigma_{gs,gt}$ denotes their covariance. $C=10^{-6}$ is used for numerical stability.


\begin{table}[t]
\caption{Zero-shot comparisons on $5$ synthetic benchmarks under integer and non-integer scales. All foundation models use their large variants. The \colorbox{best}{best} and \colorbox{best2}{second-best} results are highlighted.}\label{tab:Quantitative}
\vspace{-8pt}
\centering
\Large
\renewcommand\arraystretch{1.05}
\resizebox{\columnwidth}{!}{
\begin{tabular}{l|c|cccccccccccccc}
\specialrule{1.5pt}{0.8ex}{0.8ex} 
Methods &Datasets  &$\times2.5$ &$\times3.0$ &$\times4.0$     &$\times5.7$ &$\times6.0$ &$\times7.1$  
 &$\times8.0$ &$\times9.3$ &$\times11.0$     &$\times12.5$ &$\times13.0$ &$\times14.9$    &$\times16.0$ &$\times19.3$ \\ \midrule

PromptDA       &\multirow{11}{*}{NYU-v2}   & 17.34 & 17.06 & 17.64 & 17.57 & 17.61 & 17.58 & 18.44 & 18.01 & 18.36 & 18.56 & 18.66 & 18.49 & 19.24 & 19.00 \\
Marigold-DC   &   & 9.12 & 9.15 & 9.26 & 9.48 & 9.53 & 9.80 & 10.01 & 10.30 & 10.90 & 11.38 & 11.55 & 12.13 & 12.53 & 13.79 \\
InfiniDepth   &   & 8.91 & 8.73 & 8.70 & 8.68 & 8.71 & 9.02 & 9.15 & 9.54 & 10.15 & 10.46 & 10.86 & 11.20 & 11.65 & 13.56 \\
LDCM                &   & 4.54 & 4.57 & 4.95 & 6.15 & 6.50 & 7.04 & 7.58 & 8.44 & 9.18 & 9.89 & 10.28 & 11.12 & 11.52 & 13.12  \\
FDSR              &   & 3.90 & 3.64 & 4.03 & 3.85 & 4.08 & 4.28 & 4.99 & 5.13 & 5.60 & 6.30 & 6.82 & 9.00 & 9.98 & 12.29 \\
DCTNet         &   & 3.33 & 3.41 & 3.54 & 4.05 & 4.17 & 4.64 & 5.01 & 5.65 & 6.57 & 7.56 & 7.93 & 9.22 & 9.98 & 11.97 \\
DORNet           &   & 3.82 & 3.71 & 3.92 & 4.05 & 4.14 & 4.21 & 4.80 & 5.28 & 5.75 & 6.69 & 7.30 & 8.64 & 9.95 & 12.23 \\
C2PD               &   & 3.51 & 3.43 & 3.73 & 3.67 & 3.99 & 4.19 & 4.76 & 5.02 & 5.67 & 6.09 & 7.03 & 7.96 & 10.24 & 12.62\\
DuCos              &   & 3.38 & 3.17 & 3.33 & 3.93 & 4.04 & 4.47 & 4.80 & 5.33 & 5.91 & 6.36 & 6.64 & 7.87 & 8.80 & 11.46 \\
\textbf{FoundDSR-S}                    &   & \colorbox{best2}{2.23} & \colorbox{best2}{2.19} & \colorbox{best2}{2.31} & \colorbox{best2}{2.84} & \colorbox{best2}{2.95} & \colorbox{best2}{3.32} & \colorbox{best2}{3.60} & \colorbox{best2}{4.03} & \colorbox{best2}{4.57} & \colorbox{best2}{5.08} & \colorbox{best2}{5.33} & \colorbox{best2}{6.21} & \colorbox{best2}{6.85} & \colorbox{best2}{8.79} \\
\textbf{FoundDSR}                      &   & \colorbox{best}{1.95} & \colorbox{best}{1.95} & \colorbox{best}{2.17} & \colorbox{best}{2.71} & \colorbox{best}{2.81} & \colorbox{best}{3.16} & \colorbox{best}{3.42} & \colorbox{best}{3.84} & \colorbox{best}{4.33} & \colorbox{best}{4.75} & \colorbox{best}{4.96} & \colorbox{best}{5.70} & \colorbox{best}{6.21} & \colorbox{best}{7.91} \\

\specialrule{1pt}{0.4ex}{0.4ex}
PromptDA       &\multirow{10}{*}{Lu}   & 22.12 & 22.20 & 22.33 & 22.58 & 22.58 & 22.86 & 23.02 & 23.03 & 23.53 & 23.49 & 23.72 & 23.90 & 23.37 & 23.75 \\
Marigold-DC   &   & 36.74 & 36.76 & 36.47 & 35.82 & 36.11 & 35.01 & 36.51 & 27.42 & 37.30 & 33.88 & 34.25 & 26.18 & 36.11 & 35.47 \\
LDCM                &   & 2.49 & 2.18 & 2.73 & 3.92 & 4.19 & 4.64 & 5.14 & 5.80 & 6.21 & 6.67 & 6.94 & 7.59 & 7.65 & 8.68 \\
FDSR              &   & 2.03 & 1.78 & 2.09 & 2.07 & 2.14 & 2.49 & 3.09 & 3.34 & 4.04 & 4.70 & 5.04 & 6.21 & 6.35 & 8.01 \\
DCTNet         &   & 2.02 & 1.99 & 1.99 & 2.22 & 2.28 & 2.64 & 2.98 & 3.57 & 4.38 & 5.07 & 5.35 & 6.23 & 6.59 & 7.82 \\
DORNet           &   & 2.39 & 2.39 & 2.41 & 2.57 & 2.65 & 2.94 & 3.16 & 3.65 & 4.40 & 4.98 & 5.19 & 5.88 & 6.35 & 7.76  \\
C2PD               &   & 2.84 & 2.42 & 2.54 & 2.41 & 2.92 & 2.96 & 3.91 & 3.72 & 4.54 & 4.87 & 5.27 & 5.74 & 7.42 & 8.87 \\
DuCos              &   & 2.24 & 2.15 & 2.17 & 2.32 & 2.40 & 2.56 & 2.76 & 3.15 & 3.92 & 4.32 & 4.57 & 5.46 & 5.99 & 7.71 \\
\textbf{FoundDSR-S}                    &   & \colorbox{best2}{1.09} & \colorbox{best2}{1.03} & \colorbox{best2}{1.03} & \colorbox{best2}{1.25} & \colorbox{best2}{1.32} & \colorbox{best2}{1.49} & \colorbox{best2}{1.69} & \colorbox{best2}{1.99} & \colorbox{best2}{2.51} & \colorbox{best2}{3.02} & \colorbox{best2}{3.27} & \colorbox{best2}{4.05} & \colorbox{best2}{4.57} & \colorbox{best2}{6.00} \\
\textbf{FoundDSR}                      &   & \colorbox{best}{0.94} & \colorbox{best}{0.90} & \colorbox{best}{0.93} & \colorbox{best}{1.17} & \colorbox{best}{1.22} & \colorbox{best}{1.46} & \colorbox{best}{1.61} & \colorbox{best}{1.92} & \colorbox{best}{2.40} & \colorbox{best}{2.85} & \colorbox{best}{3.05} & \colorbox{best}{3.72} & \colorbox{best}{4.09} & \colorbox{best}{5.37} \\

\specialrule{1pt}{0.4ex}{0.4ex}
PromptDA       &\multirow{10}{*}{Middlebury}   & 14.57 & 14.70 & 14.61 & 15.06 & 14.98 & 15.24 & 15.16 & 15.23 & 15.59 & 15.79 & 15.85 & 15.49 & 16.13 & 16.22 \\
Marigold-DC   &   & 5.08 & 5.56 & 6.01 & 5.97 & 5.72 & 5.92 & 6.10 & 6.17 & 6.48 & 6.97 & 6.47 & 7.12 & 7.44 & 8.22 \\
LDCM                &   & 2.26 & 2.17 & 2.34 & 3.30 & 3.44 & 3.93 & 4.30 & 4.73 & 5.19 & 5.62 & 5.65 & 6.16 & 6.39 & 7.15 \\
FDSR              &   & 1.61 & 1.64 & 1.87 & 2.01 & 2.09 & 2.24 & 2.54 & 2.78 & 3.15 & 3.77 & 3.93 & 4.98 & 5.53 & 6.66 \\
DCTNet         &   & 1.79 & 1.82 & 1.90 & 2.22 & 2.27 & 2.56 & 2.82 & 3.22 & 3.72 & 4.36 & 4.48 & 5.21 & 5.61 & 6.63 \\
DORNet           &   & 1.99 & 1.99 & 2.01 & 2.16 & 2.19 & 2.37 & 2.58 & 2.85 & 3.18 & 3.65 & 3.72 & 4.39 & 4.76 & 6.13 \\
C2PD               &   & 1.78 & 1.86 & 1.97 & 2.01 & 2.10 & 2.21 & 2.41 & 2.64 & 3.04 & 3.30 & 3.69 & 4.35 & 5.39 & 6.81 \\
DuCos              &   & 1.87 & 1.77 & 1.81 & 1.99 & 2.05 & 2.22 & 2.40 & 2.67 & 2.97 & 3.16 & 3.38 & 3.96 & 4.54 & 6.07 \\
\textbf{FoundDSR-S}                    &   & \colorbox{best2}{1.26} & \colorbox{best2}{1.25} & \colorbox{best2}{1.31} & \colorbox{best2}{1.50} & \colorbox{best2}{1.52} & \colorbox{best2}{1.70} & \colorbox{best2}{1.85} & \colorbox{best2}{2.12} & \colorbox{best2}{2.37} & \colorbox{best2}{2.63} & \colorbox{best2}{2.76} & \colorbox{best2}{3.18} & \colorbox{best2}{3.57} & \colorbox{best2}{4.68} \\
\textbf{FoundDSR}                      &   & \colorbox{best}{0.99} & \colorbox{best}{0.98} & \colorbox{best}{1.06} & \colorbox{best}{1.26} & \colorbox{best}{1.29} & \colorbox{best}{1.48} & \colorbox{best}{1.62} & \colorbox{best}{1.87} & \colorbox{best}{2.13} & \colorbox{best}{2.41} & \colorbox{best}{2.50} & \colorbox{best}{2.95} & \colorbox{best}{3.35} & \colorbox{best}{4.37} \\

\specialrule{1pt}{0.4ex}{0.4ex}
PromptDA       &\multirow{11}{*}{RGB-D-D}   & 4.89 & 4.89 & 4.93 & 5.06 & 5.06 & 5.17 & 5.24 & 5.33 & 5.39 & 5.46 & 5.46 & 5.47 & 5.54 & 5.53 \\
Marigold-DC   &   & 3.15 & 3.19 & 3.24 & 3.38 & 3.44 & 3.58 & 3.73 & 3.94 & 4.19 & 4.49 & 4.61 & 4.95 & 5.22 & 5.75 \\
InfiniDepth   &   & 3.61 & 3.55 & 3.55 & 3.46 & 3.50 & 3.57 & 3.67 & 3.80 & 4.03 & 4.22 & 4.39 & 4.57 & 4.85 & 5.60  \\
LDCM                &   & 2.14 & 2.08 & 2.40 & 2.73 & 2.82 & 3.14 & 3.54 & 3.84 & 4.05 & 4.41 & 4.51 & 4.94 & 5.28 & 5.79  \\
FDSR              &   & 1.58 & 1.68 & 1.89 & 1.90 & 1.98 & 2.08 & 2.39 & 2.57 & 2.74 & 3.06 & 3.23 & 3.96 & 4.48 & 5.24 \\
DCTNet         &   & 1.80 & 1.83 & 1.94 & 2.25 & 2.32 & 2.55 & 2.73 & 2.96 & 3.27 & 3.57 & 3.69 & 4.12 & 4.42 & 5.13 \\
DORNet           &   & 1.82 & 1.85 & 1.93 & 2.19 & 2.25 & 2.47 & 2.66 & 2.92 & 3.25 & 3.50 & 3.59 & 3.94 & 4.27 & 5.05 \\
C2PD              &   & 1.61 & 1.77 & 2.01 & 1.90 & 2.10 & 2.17 & 2.54 & 2.65 & 2.87 & 3.08 & 3.43 & 3.94 & 4.69 & 5.55 \\
DuCos              &   & 1.54 & 1.51 & 1.62 & 1.89 & 1.94 & 2.13 & 2.29 & 2.53 & 2.88 & 3.14 & 3.23 & 3.57 & 3.90 & 4.85 \\
\textbf{FoundDSR-S}                    &   & \colorbox{best2}{1.26} & \colorbox{best2}{1.25} & \colorbox{best2}{1.30} & \colorbox{best2}{1.52} & \colorbox{best2}{1.56} & \colorbox{best2}{1.74} & \colorbox{best2}{1.89} & \colorbox{best2}{2.08} & \colorbox{best2}{2.32} & \colorbox{best2}{2.55} & \colorbox{best2}{2.59} & \colorbox{best2}{2.88} & \colorbox{best2}{3.11} & \colorbox{best2}{3.78} \\
\textbf{FoundDSR}                      &   & \colorbox{best}{1.08} & \colorbox{best}{1.08} & \colorbox{best}{1.17} & \colorbox{best}{1.41} & \colorbox{best}{1.46} & \colorbox{best}{1.63} & \colorbox{best}{1.78} & \colorbox{best}{1.98} & \colorbox{best}{2.21} & \colorbox{best}{2.43} & \colorbox{best}{2.46} & \colorbox{best}{2.71} & \colorbox{best}{2.90} & \colorbox{best}{3.43} \\

\specialrule{1pt}{0.4ex}{0.4ex}
PromptDA       &\multirow{10}{*}{iBims-1}   & 37.84 & 38.21 & 38.40 & 38.93 & 39.39 & 38.70 & 40.08 & 40.95 & 41.85 & 42.65 & 43.06 & 43.12 & 44.17 & 45.10 \\
Marigold-DC  &   & 39.83 & 40.27 & 37.83 & 37.04 & 37.08 & 37.12 & 36.74 & 35.83 & 37.45 & 38.33 & 38.27 & 38.17 & 39.65 & 41.46 \\
LDCM                &   & 23.10 & 21.65 & 21.93 & 24.28 & 25.02 & 26.06 & 27.13 & 29.18 & 30.31 & 31.73 & 32.37 & 33.81 & 34.53 & 37.18 \\
FDSR            &   & 14.82 & 15.63 & 16.67 & 19.15 & 19.41 & 20.79 & 21.83 & 23.20 & 25.06 & 27.23 & 28.16 & 31.07 & 32.91 & 36.10  \\
DCTNet         &   & 13.80 & 14.26 & 16.15 & 19.36 & 19.86 & 21.62 & 22.67 & 24.37 & 26.25 & 28.11 & 28.80 & 30.95 & 32.32 & 35.64 \\
DORNet           &   & 13.01 & 13.61 & 15.36 & 18.70 & 19.23 & 21.09 & 22.20 & 24.20 & 26.14 & 27.90 & 28.59 & 30.64 & 32.10 & 35.46  \\
C2PD               &   & 12.38 & 14.02 & 14.87 & 16.93 & 18.05 & 19.54 & 20.80 & 22.22 & 24.77 & 26.42 & 27.97 & 29.57 & 33.11 & 36.94  \\
DuCos            &   & 13.29 & 13.23 & 15.27 & 18.05 & 18.51 & 20.31 & 21.48 & 23.43 & 25.31 & 26.98 & 27.74 & 29.95 & 31.57 & 35.38 \\
\textbf{FoundDSR-S}                    &   & \colorbox{best2}{12.80} & \colorbox{best2}{12.76} & \colorbox{best2}{13.47} & \colorbox{best2}{15.76} & \colorbox{best2}{16.18} & \colorbox{best2}{17.72} & \colorbox{best2}{18.84} & \colorbox{best2}{20.72} & \colorbox{best2}{22.94} & \colorbox{best2}{24.81} & \colorbox{best2}{25.51} & \colorbox{best2}{27.28} & \colorbox{best2}{28.55} & \colorbox{best2}{31.77} \\
\textbf{FoundDSR}                      &   & \colorbox{best}{11.19} & \colorbox{best}{11.33} & \colorbox{best}{12.52} & \colorbox{best}{15.11} & \colorbox{best}{15.58} & \colorbox{best}{17.03} & \colorbox{best}{18.19} & \colorbox{best}{20.10} & \colorbox{best}{22.21} & \colorbox{best}{24.00} & \colorbox{best}{24.70} & \colorbox{best}{26.48} & \colorbox{best}{27.70} & \colorbox{best}{30.57} \\

\specialrule{1.5pt}{0.8ex}{0.8ex}
\end{tabular}}
\vspace{-19pt}
\end{table}

\section{Experiments}
\label{sec:exp}

\subsection{Experimental Setups}
\noindent \textbf{Datasets.} 
To train FoundDSR, we construct a large-scale dataset of approximately $1.4$ million RGB-D pairs from DIML~\cite{cho2021deep, kim2016structure, kim2017deep, kim2018deep}, Hypersim~\cite{roberts2021hypersim}, DynamicReplica~\cite{karaev2023dynamicstereo}, and LingBot-Depth~\cite{tan2026masked}. The trained model is directly evaluated on common benchmarks without any fine-tuning,  including $449$ pairs NYU-v2~\cite{silberman2012indoor}, $6$ pairs Lu~\cite{lu2014depth}, $30$ pairs Middlebury~\cite{hirschmuller2007evaluation,scharstein2007learning}, $100$ pairs iBims-1~\cite{koch2018evaluation}, $405$ pairs RGB-D-D~\cite{he2021towards}, and $560$ pairs TOFDSR~\cite{yan2025tri}. For synthetic data, we follow the standard protocol in prior DSR works~\cite{zhao2022discrete,wang2025dornet} that synthesize LR depth from GT depth using bicubic downsampling. For real-world scenarios, the RGB-D-D~\cite{he2021towards} and TOFDSR~\cite{yan2025tri} datasets also provide real-world LR depth captured by ToF cameras, which is subject to inherently more complex degradations than synthetic counterparts.


\noindent \textbf{Implementation Details.} 
Following prior works~\cite{kim2021deformable,zhao2022discrete}, we use root mean square error (RMSE) in centimeters as the default evaluation metric, \textit{with additional metrics reported in the appendix}. During training, FoundDSR is optimized using AdamW~\cite{loshchilov2017decoupled}, with an initial learning rate of $5 \times 10^{-6}$ for the encoder and $5 \times 10^{-5}$ for other layers. For data augmentation, we first apply random cropping, random horizontal and vertical flipping, and random 90-degree rotations. Then, we randomly sample scaling factors from $[3,11]$, providing diverse scale supervision while leaving larger factors for evaluating extrapolation ability. Gaussian noise (with zero mean and a standard deviation of $0.07$) is injected into a randomly selected $1\%$ of the training samples. The entire training process is conducted on $4$ NVIDIA RTX 4090 GPUs.

\subsection{Comparison with the State-of-the-Art}
We compare FoundDSR with advanced DSR methods, including FDSR~\cite{he2021towards}, DCTNet~\cite{zhao2022discrete}, DORNet~\cite{wang2025dornet}, C2PD~\cite{kang2025c2pd}, and DuCos~\cite{yan2025ducos}, as well as depth foundation models using low-quality depth inputs, such as PromptDA~\cite{lin2025prompting}, Marigold-DC~\cite{viola2025marigold}, InfiniDepth~\cite{yu2026infinidepth}, and LDCM~\cite{yu2026large}.
Following prior works~\cite{lin2025prompting,yu2026large}, foundation models are evaluated using their released weights, while DSR methods are retrained on the same data as FoundDSR.

%

\noindent \textbf{Comparisons on Synthetic Dataset.} 
Tab.~\ref{tab:Quantitative} presents zero-shot comparisons over $14$ integer and non-integer scales. The results confirm that our method exhibits strong zero-shot generalization. Although recent foundation models also use low-quality depth as a prompt, they primarily regress depth from RGB rather than learning the LR-to-HR depth mapping. Consequently, FoundDSR achieves the lowest RMSE, outperforming suboptimal LDCM by $46.09\%$, $46.54\%$, $47.57\%$, $45.08\%$, and $19.78\%$ on the NYU-v2, Lu, Middlebury, RGB-D-D, and iBims-1, respectively ($\times16.0$). On the low-light Lu, Marigold-DC degrades substantially due to its strong reliance on RGB inherited from its depth estimation backbone, which become unreliable and deviate from the pretraining distribution under poor illumination. Moreover, compared with advanced DSR methods, FoundDSR shows increasing gains at larger scales, reducing the average  RMSE over DCTNet, DORNet, C2PD, and DuCos by $29.38\%$ and $36.26\%$ on NYU-v2 at $\times8.0$ and $\times16.0$, respectively. 

\begin{figure}[t]
\centering
\includegraphics[width=0.92\columnwidth]{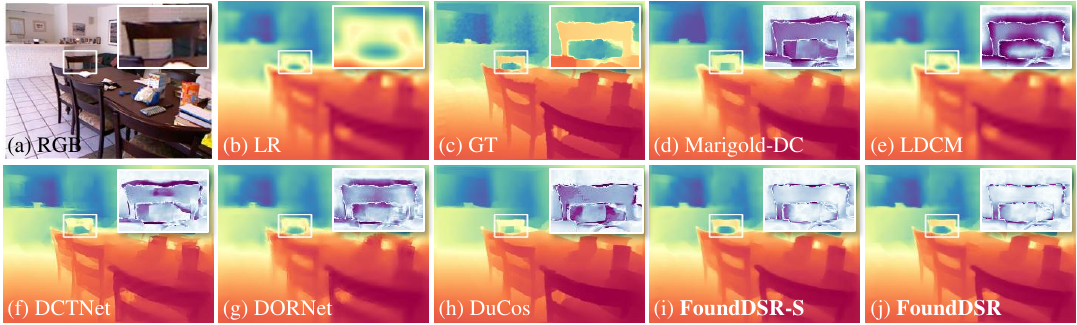}\\
\vspace{-6pt}
\caption{Visual comparisons on the $\times16.0$ NYU‑v2 dataset. The zoomed‑in patches in (d)–(j) highlight the error maps within the white boxes, where darker purple shades indicate higher errors.}\label{fig:NYU_x16_408}
\vspace{-6pt}
\end{figure}

\begin{table*}[t]
\caption{Zero-shot comparisons on the real-world RGB-D-D and TOFDSR datasets.}\label{tab:real}
    \vspace{-10pt}
	\centering
	\LARGE
	\resizebox{1\linewidth}{!}{
\begin{tabular}{c|ccccccccccc}
\specialrule{1.5pt}{0.8ex}{0.8ex} 
RMSE    &PromptDA &Marigold-DC &InfiniDepth  &LDCM  &FDSR   &DCTNet 	  &DORNet  &C2PD		&DuCos	&\textbf{FoundDSR-S}  &\textbf{FoundDSR}\\
\midrule
RGB-D-D     &7.06    &\colorbox{best2}{6.85}    &6.87    &7.65    &7.41    &7.82        &7.49    &7.26    &7.44             &6.93 &\colorbox{best}{6.68}	\\
TOFDSR     	&8.79    &7.80    &8.48    &8.22    &7.23    &7.45    &7.34       &7.23	 &7.42 &\colorbox{best2}{7.05}	          &\colorbox{best}{6.92} \\
\specialrule{1.5pt}{0.8ex}{0.8ex}
\end{tabular}}
\vspace{-10pt}
\end{table*}

\begin{table*}[!ht]
\caption{Zero-shot comparisons on multiple $\times4.0$ benchmarks with noise levels from $0.02$ to $0.04$.}\label{tab:noise}
\vspace{-10pt}
\centering
\Large
\renewcommand\arraystretch{1.05}
\resizebox{1\linewidth}{!}{
\begin{tabular}{l|ccccccccccccccc}
\specialrule{1.5pt}{0.8ex}{0.8ex} 
\multirow{2}{*}{Methods}  &\multicolumn{3}{c}{NYU-v2}   &\multicolumn{3}{c}{Lu}  &\multicolumn{3}{c}{Middlebury} &\multicolumn{3}{c}{RGB-D-D} &\multicolumn{3}{c}{iBims-1} \\ 
\cmidrule(lr){2-4}\cmidrule(lr){5-7}\cmidrule(lr){8-10}\cmidrule(lr){11-13}\cmidrule(lr){14-16}
 &0.02 &0.03 &0.04     &0.02 &0.03 &0.04      &0.02 &0.03 &0.04  &0.02 &0.03 &0.04  &0.02 &0.03 &0.04   \\ \midrule

PromptDA                   &17.42  &17.34  &17.36     &22.49  &22.42  &22.40           &15.12  &15.52  &11.93         &5.23  &5.35  &5.53  &38.83  &39.06  &39.48   \\
Marigold-DC               &8.71  &8.79  &8.97        &36.42  &36.04  &35.94         &5.75  &5.83  &5.65               &2.93  &2.93  &2.96  &33.90  &33.80  &33.89   \\
LDCM                            &6.00  &7.40  &9.24        &4.64  &6.24  &7.99            &4.33  &5.88  &7.46              &3.00  &3.62  &4.34  &23.76  &26.85  &30.44   \\
FDSR                          &5.82  &6.56  &7.05        &3.18  &3.66  &3.96           &3.00  &3.49  &3.86               &2.26  &2.56  &2.92  &18.61  &19.71  &20.62   \\
DCTNet                     &6.24  &6.94  &7.81        &3.32  &3.94  &4.30           &3.35  &3.96  &4.22               &2.58  &2.95  &3.44  &20.08  &21.83  &23.36   \\
DORNet                       &5.79  &6.81  &7.96        &3.95  &4.69  &5.45           &3.49  &4.13  &4.79               &2.65  &3.01  &3.36  &19.56  &21.67  &23.79   \\
C2PD                           &5.53  &6.42  &7.22        &2.97  &3.59  &4.08           &3.17  &3.38  &3.91               &2.40  &2.74  &3.15  &17.82  &18.94  &20.17   \\
DuCos                          &5.37  &6.61  &6.98        &2.88  &3.34  &4.12           &3.06  &3.33  &3.84               &2.17  &2.53  &2.81  &17.89  &19.57  &20.10   \\
\textbf{FoundDSR-S}                                &\colorbox{best2}{4.44}  &\colorbox{best2}{5.13}  &\colorbox{best2}{5.73}        &\colorbox{best2}{2.41}  &\colorbox{best2}{2.69}  &\colorbox{best2}{3.15}           &\colorbox{best2}{2.46}  &\colorbox{best2}{2.93}  &\colorbox{best2}{3.26}               &\colorbox{best2}{1.88}  &\colorbox{best2}{2.19}  &\colorbox{best2}{2.44}  &\colorbox{best2}{16.52}  &\colorbox{best2}{17.77}  &\colorbox{best2}{18.94}   \\
\textbf{FoundDSR}                                  &\colorbox{best}{4.21}  &\colorbox{best}{4.91}  &\colorbox{best}{5.48}        &\colorbox{best}{2.12}  &\colorbox{best}{2.59}  &\colorbox{best}{3.07}           &\colorbox{best}{2.28}  &\colorbox{best}{2.79}  &\colorbox{best}{3.20}               &\colorbox{best}{1.85}  &\colorbox{best}{2.15}  &\colorbox{best}{2.40}  &\colorbox{best}{16.26}  &\colorbox{best}{17.58}  &\colorbox{best}{18.63}   \\
\specialrule{1.5pt}{0.8ex}{0.8ex}
\end{tabular}}
 \vspace{-8pt}
\end{table*}

\begin{wrapfigure}{r}{0.44\textwidth}
\centering
\includegraphics[width=\linewidth]{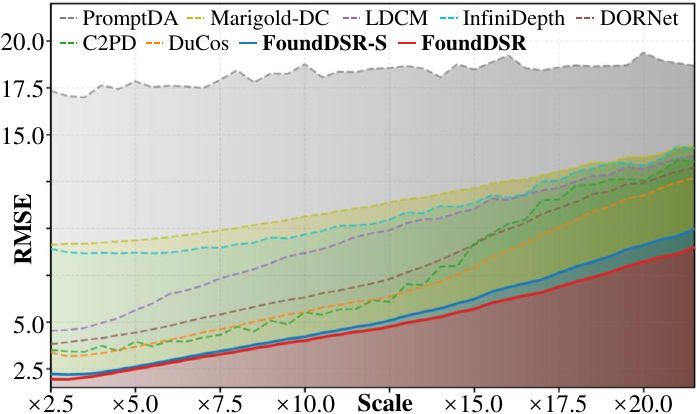}
\setlength{\abovecaptionskip}{-4pt}
\caption{Zero-shot comparisons of continuous upsampling factors with prior foundation models and DSR methods on NYU-v2.}
\label{fig:ArScales}
\vspace{-16pt}
\end{wrapfigure}
Fig.~\ref{fig:NYU_x16_408} shows the visual results, confirming that our method effectively restores degraded depth structures while preserving fine geometric information. Furthermore, Fig.~\ref{fig:ArScales} presents zero-shot comparisons across arbitrary upsampling factors. As the scale increases, all competing methods exhibit higher RMSE, whereas our FoundDSR consistently achieves the lowest error with the most gradual performance degradation, demonstrating strong generalization. 

\noindent \textbf{Comparisons on Real-World Dataset.} 
Tab.~\ref{tab:real} reports the comparisons on real-world RGB-D-D and TOFDSR datasets with unknown degradations. The results show that FoundDSR exhibits superior zero-shot generalization over both depth foundation models and existing DSR methods, outperforming the second-best method by $0.17cm$ and $0.31cm$ in RMSE on RGB-D-D and TOFDSR, respectively. \textit{Please see our appendix for visual results.}

\noindent \textbf{Robustness to Noise.} 
Following prior works~\cite{kim2021deformable, wang2025dornet}, we additionally corrupt the LR depth with Gaussian noise to simulate more challenging degradation conditions. As reported in Tab.~\ref{tab:noise}, we evaluate all methods on five benchmarks under zero-mean Gaussian noise with standard deviations of $0.02$, $0.03$, and $0.04$. FoundDSR consistently shows strong robustness across all noise levels,  reducing the average RMSE over the second-best DuCos by $14.82\%$, $15.15\%$, and $13.39\%$ across all datasets at noise levels of $0.02$, $0.03$, and $0.04$, respectively. \textit{Qualitative comparisons are provided in the appendix.}

\begin{table*}[t]
\centering
\begin{minipage}[t]{0.475\linewidth}
    \centering
    \caption{Complexity on $\times16.0$ NYU-v2, evaluated on a single NVIDIA RTX 4090 GPU.}\label{tab:complex}
    \vspace{-8pt}
    \setlength{\tabcolsep}{2pt} 
    \resizebox{\linewidth}{!}{
    \begin{tabular}{l|cccc|c}
    \specialrule{1.5pt}{0.8ex}{0.8ex} 
    Methods    & \makecell[c]{Param. \\(M)} & \makecell[c]{Memo. \\(G)} & \makecell[c]{FLOPs \\(G)}  &\makecell[c]{Time \\(s)}   & RMSE\\
    \midrule
    C2PD                  &62.05    &\colorbox{best2}{1.75}    &\colorbox{best2}{414.75}    &0.061    &8.80    	\\
    DuCos                 &\colorbox{best}{34.38}    &9.02    &2927.49    &\colorbox{best}{0.025}    &10.24    	\\
    Marigold-DC           &949.58    &9.83    &94753.76    &14.02    &13.79    	\\
    InfiniDepth           &358.15    &23.11    &16446.19    &0.081    &13.56    	\\
    LDCM                  &238.62    &3.01    &903.07    &0.069    &13.12    	\\
    \textbf{FoundDSR-S}   &\colorbox{best2}{51.85}    &\colorbox{best}{1.08}    &\colorbox{best}{90.76}    &\colorbox{best}{0.016}    &\colorbox{best2}{6.85}    	\\
    \textbf{FoundDSR}     &657.44    &3.65    &1248.86    &0.027    &\colorbox{best}{6.21}    	\\
    \specialrule{1.5pt}{0.8ex}{0.8ex}
    \end{tabular}}
\end{minipage}
\hfill
\begin{minipage}[t]{0.48\linewidth}
    \centering
    \caption{Ablation study of position (Pos.) and covariance (Cov.) modulation in our guided 2D GS strategy on $\times13.0$, $\times14.9$, and $\times16.0$ NYU-v2, with time measured at the $\times16.0$ scale.}\label{tab:abl_2dgs}
    \vspace{-8pt}
    \setlength{\tabcolsep}{2pt}  
    \Large 
    \resizebox{\linewidth}{!}{
    \begin{tabular}{l|ccc|cc|ccc}
    \specialrule{1.5pt}{0.8ex}{0.8ex} 
                    &2D GS &Pos. &Cov.  & \makecell[c]{Param. \\(M)} &\makecell[c]{Time \\(s)}   &$\times13.0$ &$\times14.9$  &$\times16.0$ \\
    \midrule
    (a)             &    &    &                                  &\colorbox{best}{48.52}    &\colorbox{best}{0.013}     &6.75    &8.48    &9.42   	\\
    (b)             &\checkmark    &    &                        &\colorbox{best2}{51.04}    &\colorbox{best2}{0.015}     &5.63     &6.56    &7.26   	\\
    (c)             &\checkmark    &\checkmark    &              &51.85    &0.016     &5.60     &6.50    &7.18  	\\
    (d)             &\checkmark    &    &\checkmark              &51.85    &0.016     &\colorbox{best2}{5.52}    &\colorbox{best2}{6.39}    &\colorbox{best2}{7.02}   	\\
    (e)             &\checkmark    &\checkmark    &\checkmark    &51.85    &0.016     &\colorbox{best}{5.33}    &\colorbox{best}{6.21}    &\colorbox{best}{6.85}   	\\
    \specialrule{1.5pt}{0.8ex}{0.8ex}
    \end{tabular}}
\end{minipage}
\vspace{-16pt}
\end{table*}

\noindent \textbf{Complexity Analysis.} 
Tab.~\ref{tab:complex} compares model complexity and computational efficiency.  As a foundation model, FoundDSR inevitably involves a relatively large number of parameters. Nevertheless, our lightweight FoundDSR-S remains competitive with existing DSR methods such as C2PD and DuCos in parameter count and GPU memory usage, while requiring fewer FLOPs, lower inference latency, and substantially improving reconstruction accuracy. Compared with previous foundation models, FoundDSR achieves a more favorable trade-off between computational cost and reconstruction accuracy. Notably, Marigold-DC is built upon a latent diffusion model and requires a $50$-step denoising process, resulting in significantly higher inference costs than other approaches. For consistency with other methods, we report only the FLOPs of a forward pass for Marigold-DC.  Compared with the second-best foundation model LDCM, FoundDSR reduces inference time by $60.87\%$ while improving reconstruction performance by $52.67\%$.

\subsection{Ablation Study}

\noindent \textbf{Guided 2D Gaussian Splatting.} 
Tab.~\ref{tab:abl_2dgs} presents the ablation study of our guided 2D GS, where baseline (a) removes the entire guided 2D GS. \textit{For efficiency, all ablations are conducted using our ViT-S variant.} Compared with (a), standard 2D GS (b) consistently improves performance across different scales, validating the effectiveness of continuous Gaussian representations for arbitrary-scale depth upsampling. Building upon (b), (c) and (d) further enhance reconstruction quality by incorporating our RGB-guided strategy to modulate the positions or covariances of Gaussian primitives, with only marginal computational overhead.  When both are jointly modulated, our method (e) achieves the best performance, reducing RMSE over baseline (a) by $1.42cm$, $2.27cm$, and $2.57cm$ on NYU-v2 at $\times13.0$, $\times14.9$, and $\times16.0$, respectively.

\begin{wrapfigure}{r}{0.49\textwidth}
\vspace{-10pt}
\centering
\includegraphics[width=\linewidth]{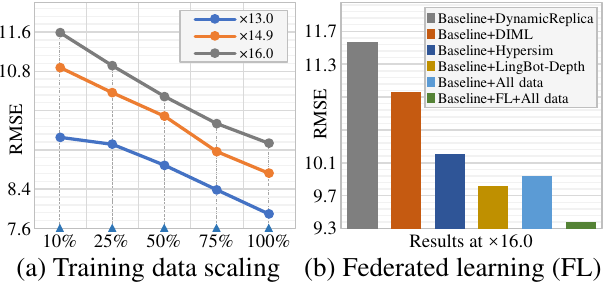}
\setlength{\abovecaptionskip}{-8pt}
\caption{Ablation study on (a) training data scaling and (b) heterogeneous federated learning. RMSE is averaged over five synthetic datasets.}
\label{fig:scaling_fl}
\vspace{-10pt}
\end{wrapfigure}

\noindent \textbf{Training Data Scaling.} 
To evaluate the scalability of our method, we construct training subsets containing $10\%$ to $100\%$ of the full dataset by random sampling, while maintaining the same sampling ratio across all data sources. As shown in Fig.~\ref{fig:scaling_fl}(a), increasing the data consistently improves performance across different upsampling factors, demonstrating the strong data-scaling capability of FoundDSR. Compared with using only $10\%$ of the data, training on the full dataset reduces RMSE by $1.56cm$, $2.15cm$, and $2.25cm$ at $\times13.0$, $\times14.9$, and $\times16.0$, respectively.

\noindent \textbf{Federated Learning.} 
Fig.~\ref{fig:scaling_fl}(b) presents the ablation study of federated learning, where the baseline removes all federated learning components. These results indicate that directly mixing all data sources fails to achieve proportional gains from the increased data scale and even underperforms single-source training on LingBot-Depth, highlighting the optimization interference caused by distribution discrepancies.  With our heterogeneous federated learning strategy, our method effectively alleviates such interference and achieves the best performance. Specifically, it reduces the average RMSE by $0.56cm$ compared with direct mixed-data training. These quantitative comparisons demonstrate the scalability of our method to larger-scale and more diverse data sources.

\section{Conclusion}
In this paper, we propose FoundDSR, the first foundation model for depth super-resolution, trained on million-scale heterogeneous data to substantially enhance zero-shot generalization to out-of-distribution data and arbitrary upsampling factors. To facilitate continuous and arbitrary-scale depth upsampling, we introduce an RGB-guided 2D Gaussian Splatting that leverages high-resolution RGB prompts to modulate the position and covariance of each Gaussian primitive, enabling anisotropic deformation in position, shape, and orientation along structural directions. Furthermore, FoundDSR incorporates heterogeneous federated learning that effectively mitigates optimization instability and dominant-source bias caused by distribution gaps across large-scale non-uniform data sources, thereby improving scalability to larger and more diverse training data. Extensive zero-shot experiments under synthetic, real-world, arbitrary-scale, and noisy settings consistently demonstrate the strong generalization of FoundDSR to out-of-distribution data.

\subsection*{AI use statement}

In this work, we used generative AI tools for translation, language polishing, expression refinement, and the detection of typographical and grammatical errors. We have reviewed all AI-assisted work. We take responsibility for the final content of this work, including text, claims, or artifacts produced with the aid of generative AI.





\subsection*{Ethics statement}

Our study focuses on depth super-resolution, a fundamental problem in computer vision. The training and evaluation are conducted using publicly available RGB-D datasets from diverse sources, without collecting or processing personally identifiable or sensitive information. This research does not involve human subjects and has been conducted in accordance with standard research ethics and responsible research practices.  


\subsection*{Reproducibility statement}

To facilitate verification and extension of our work, we provide detailed descriptions of the model architecture, training settings, and evaluation protocols in the paper and supplementary materials. The complete source code and trained model weights will be made publicly available after publication to support reproducibility and further research.




\bibliography{iclr2027_conference}
\bibliographystyle{iclr2027_conference}

\newpage
\appendix
\section{Appendix}

\subsection{Dataset Details}
\begin{wraptable}{r}{0.52\columnwidth}   %
\vspace{-12pt}
\caption{Overview of the training datasets.}\label{tab:datasets}
\centering
\resizebox{\linewidth}{!}{
\begin{tabular}{lccc}
\specialrule{1.5pt}{0.8ex}{0.8ex} 
Dataset    & Type        & Resolution       & Statistic \\
\midrule
DIML             & Real-World & $1344\times756$  & 1.6K   \\
Hypersim         & Synthetic  & $1024\times768$  & 46.1K  \\
DynamicReplica   & Synthetic  & $1280\times720$  & 336.9K \\
LingBot-Depth    & Synthetic  & $1280\times960$  & 1.0M   \\
\midrule
Total            & -          & -                & 1.4M   \\
\specialrule{1.5pt}{0.8ex}{0.8ex}
\end{tabular}}
\vspace{-8pt}
\end{wraptable}
As shown in Tab.~\ref{tab:datasets}, we collect $4$ publicly available RGB-D datasets to train our FoundDSR, consisting of  $1$ real-world and $3$ synthetic datasets, covering diverse scenes.

\textbf{DIML Dataset.} DIML~\cite{cho2021deep, kim2016structure, kim2017deep, kim2018deep} is captured in real-world environments using Kinect v2 and ZED stereo cameras. We use only its indoor subset, which contains approximately $1.6$K training samples spanning 18 indoor scenes.

\textbf{Hypersim Dataset.} Hypersim~\cite{roberts2021hypersim} is generated from a large repository of synthetic scenes created by professional artists, containing $77.4$K samples across $461$ indoor scenes. We remove RGB-D pairs with invalid depth values, leaving approximately $46.1$K samples for training.

\textbf{DynamicReplica Dataset.} DynamicReplica~\cite{karaev2023dynamicstereo} is built upon Replica~\cite{replica19arxiv} by introducing moving humans and animals into scanned indoor environments, yielding richer and more diverse dynamic scenes. It contains $524$ synthetic video sequences.

\textbf{LingBot-Depth Dataset.} LingBot-Depth~\cite{tan2026masked} contains large-scale real-world and synthetic RGB-D data. Since the real-world subset includes a considerable number of invalid depth samples, we only adopt the synthetic subset. It simulates real-world sensing imperfections and renders paired RGB-D data in Blender using self-hosted 3D assets, covering $442$ indoor scenes.

\subsection{Metrics}
In the main paper, we follow common practice and adopt root mean square error (RMSE) as the primary evaluation metric. For a more comprehensive evaluation, additional quantitative comparisons using mean absolute error (MAE) and ${\delta}_{1.05}$ are provided in the supplementary material. These metrics are defined as follows:  
\begin{equation}
\begin{split}
&\text{RMSE}=\sqrt{\frac{1}{|\mathbb{Q}|}\sum\limits{{{\left( \boldsymbol D_{gt}^{q}- \boldsymbol D_{hr}^{q} \right)}}}^2}, \\
&\text{MAE}=\frac{1}{|\mathbb{Q}|}\sum\limits{{{\left| \boldsymbol D_{gt}^{q}- \boldsymbol D_{hr}^{q} \right|}}}, \\
&\delta_{1.05}=\frac{|\mathbb{S} |}{|\mathbb{Q} |} ,\mathbb{S}:max(\frac{\boldsymbol D_{gt}^{q}}{\boldsymbol D_{hr}^{q}},\frac{\boldsymbol D_{hr}^{q}}{\boldsymbol D_{gt}^{q}} )<1.05,
\end{split}
\end{equation}
where $\boldsymbol D_{gt}$ and $\boldsymbol D_{hr}$ denote the ground-truth depth and the predicted HR depth, respectively. $\mathbb{Q}$ is the valid pixel set of $\boldsymbol D_{gt}$, and $q$ is one of the valid pixels.


\begin{table}[t]
\caption{Quantitative comparisons on five synthetic benchmarks using the \textbf{MAE}$\downarrow$ (cm) metric.}\label{tab:Supp_MAE}
\vspace{-4pt}
\centering
\Large
\renewcommand\arraystretch{1.05}
\resizebox{\columnwidth}{!}{
\begin{tabular}{l|c|cccccccccccccc}
\specialrule{1.5pt}{0.8ex}{0.8ex} 
Methods &Datasets  &$\times2.5$ &$\times3.0$ &$\times4.0$     &$\times5.7$ &$\times6.0$ &$\times7.1$  
 &$\times8.0$ &$\times9.3$ &$\times11.0$     &$\times12.5$ &$\times13.0$ &$\times14.9$    &$\times16.0$ &$\times19.3$ \\ \midrule

PromptDA       &\multirow{11}{*}{NYU-v2}   & 7.58 & 7.54 & 7.67 & 7.74 & 7.73 & 7.67 & 8.04 & 7.82 & 8.01 & 9.16 & 8.16 & 8.23 & 8.50 & 8.49 \\
Marigold-DC   &                            & 4.18 & 4.20 & 4.27 & 4.39 & 4.41 & 4.56 & 4.68 & 4.83 & 5.14 & 5.38 & 5.53 & 5.79 & 6.04 & 6.73 \\
InfiniDepth   &                            & 3.33 & 3.37 & 3.42 & 3.53 & 3.59 & 3.78 & 3.91 & 4.60 & 4.52 & 4.72 & 4.94 & 5.11 & 5.45 & 6.53 \\
LDCM          &                            & 1.57 & 1.63 & 1.82 & 2.32 & 2.47 & 2.64 & 3.01 & 3.40 & 3.86 & 4.25 & 4.51 & 4.95 & 5.24 & 6.19 \\
FDSR          &                            & 1.12 & 1.10 & 1.19 & 1.28 & 1.33 & 1.48 & 1.68 & 1.84 & 2.10 & 2.49 & 2.74 & 3.70 & 4.34 & 5.54 \\
DCTNet        &                            & 1.26 & 1.29 & 1.36 & 1.58 & 1.63 & 1.82 & 2.01 & 2.30 & 2.84 & 3.36 & 3.56 & 4.17 & 4.62 & 5.65 \\
DORNet        &                            & 1.07 & 1.09 & 1.11 & 1.19 & 1.23 & 1.33 & 1.54 & 1.75 & 1.86 & 2.42 & 2.75 & 3.47 & 4.11 & 5.38 \\
C2PD          &                            & 1.14 & 1.16 & 1.21 & 1.29 & 1.37 & 1.50 & 1.63 & 1.82 & 2.15 & 2.39 & 2.82 & 3.20 & 4.26 & 5.78 \\
DuCos         &                            & 0.86 & 0.83 & 0.93 & 1.19 & 1.24 & 1.43 & 1.57 & 1.79 & 2.08 & 2.36 & 2.51 & 3.18 & 3.73 & 5.17 \\
\textbf{FoundDSR-S}                    &   & \colorbox{best2}{0.84} & \colorbox{best2}{0.82} & \colorbox{best2}{0.87} & \colorbox{best2}{1.07} & \colorbox{best2}{1.11} & \colorbox{best2}{1.26} & \colorbox{best2}{1.38} & \colorbox{best2}{1.56} & \colorbox{best2}{1.81} & \colorbox{best2}{2.09} & \colorbox{best2}{2.22} & \colorbox{best2}{2.65} & \colorbox{best2}{3.02} & \colorbox{best2}{4.00} \\
\textbf{FoundDSR}                      &   & \colorbox{best}{0.81} & \colorbox{best}{0.79} & \colorbox{best}{0.84} & \colorbox{best}{1.04} & \colorbox{best}{1.08} & \colorbox{best}{1.23} & \colorbox{best}{1.34} & \colorbox{best}{1.51} & \colorbox{best}{1.73} & \colorbox{best}{1.96} & \colorbox{best}{2.07} & \colorbox{best}{2.44} & \colorbox{best}{2.74} & \colorbox{best}{3.59} \\

\specialrule{1pt}{0.4ex}{0.4ex}
Marigold-DC   &\multirow{9}{*}{Lu}         & 11.59 & 11.69 & 11.53 & 11.51 & 11.55 & 11.28 & 11.73 & 8.62 & 11.78 & 11.34 & 11.74 & 8.90 & 12.23 & 12.65 \\
LDCM          &                            & 0.77 & 0.77 & 0.91 & 1.26 & 1.37 & 1.54 & 1.68 & 1.98 & 2.19 & 2.44 & 2.56 & 2.91 & 2.97 & 3.56 \\
FDSR          &                            & 0.60 & 0.55 & 0.63 & 0.70 & 0.72 & 0.80 & 0.91 & 1.03 & 1.21 & 1.45 & 1.58 & 2.10 & 2.37 & 3.11 \\
DCTNet        &                            & 0.77 & 0.79 & 0.79 & 0.88 & 0.90 & 1.01 & 1.10 & 1.31 & 1.60 & 1.89 & 1.98 & 2.41 & 2.58 & 3.22 \\
DORNet        &                            & 0.57 & 0.58 & 0.59 & 0.68 & 0.70 & 0.80 & 0.88 & 1.02 & 1.25 & 1.46 & 1.54 & 1.89 & 2.12 & 2.92 \\
C2PD          &                            & 0.73 & 0.65 & 0.74 & 0.71 & 0.79 & 0.84 & 0.98 & 1.05 & 1.27 & 1.40 & 1.55 & 1.76 & 2.35 & 3.41 \\
DuCos         &                            & 0.51 & 0.50 & 0.52 & 0.59 & 0.62 & 0.68 & 0.75 & \colorbox{best2}{0.86} & 1.06 & 1.21 & 1.29 & 1.72 & 1.99 & 2.89 \\
\textbf{FoundDSR-S}                    &   & \colorbox{best}{0.45} & \colorbox{best}{0.43} & \colorbox{best}{0.44} & \colorbox{best}{0.49} & \colorbox{best}{0.51} & \colorbox{best}{0.56} & \colorbox{best}{0.61} & \colorbox{best}{0.71} & \colorbox{best2}{0.84} & \colorbox{best2}{0.99} & \colorbox{best2}{1.05} & \colorbox{best2}{1.34} & \colorbox{best2}{1.53} & \colorbox{best2}{2.18} \\
\textbf{FoundDSR}                      &   & \colorbox{best2}{0.49} & \colorbox{best2}{0.48} & \colorbox{best2}{0.48} & \colorbox{best2}{0.53} & \colorbox{best2}{0.55} & \colorbox{best2}{0.59} & \colorbox{best2}{0.64} & \colorbox{best}{0.71} & \colorbox{best}{0.83} & \colorbox{best}{0.96} & \colorbox{best}{1.00} & \colorbox{best}{1.24} & \colorbox{best}{1.37} & \colorbox{best}{1.93} \\

\specialrule{1pt}{0.4ex}{0.4ex}
Marigold-DC   &\multirow{9}{*}{Middlebury} & 2.33 & 2.41 & 2.54 & 2.65 & 2.59 & 2.73 & 2.78 & 2.93 & 3.12 & 3.38 & 3.28 & 3.66 & 3.80 & 4.33 \\
LDCM          &                            & 0.93 & 0.95 & 1.08 & 1.45 & 1.51 & 1.74 & 1.93 & 2.17 & 2.47 & 2.75 & 2.80 & 3.15 & 3.31 & 3.86 \\
FDSR          &                            & 0.76 & 0.78 & 0.86 & 0.94 & 0.97 & 1.04 & 1.14 & 1.24 & 1.42 & 1.71 & 1.81 & 2.43 & 2.77 & 3.55 \\
DCTNet        &                            & 0.93 & 0.95 & 0.98 & 1.09 & 1.11 & 1.23 & 1.35 & 1.55 & 1.85 & 2.20 & 2.28 & 2.73 & 2.97 & 3.64 \\
DORNet        &                            & 0.80 & 0.81 & 0.83 & 0.91 & 0.93 & 1.01 & 1.09 & 1.21 & 1.38 & 1.61 & 1.68 & 2.11 & 2.38 & 3.26 \\
C2PD          &                            & 0.89 & 0.92 & 0.94 & 0.95 & 0.98 & 1.03 & 1.10 & 1.18 & 1.35 & 1.47 & 1.64 & 1.97 & 2.55 & 3.63 \\
DuCos         &                            & 0.73 & 0.71 & 0.74 & 0.84 & 0.86 & 0.94 & 1.01 & 1.12 & 1.25 & 1.37 & 1.46 & 1.84 & 2.17 & 3.14 \\
\textbf{FoundDSR-S}                    &   & \colorbox{best2}{0.66} & \colorbox{best2}{0.65} & \colorbox{best2}{0.67} & \colorbox{best2}{0.75} & \colorbox{best2}{0.76} & \colorbox{best2}{0.84} & \colorbox{best2}{0.91} & \colorbox{best2}{1.01} & \colorbox{best2}{1.12} & \colorbox{best2}{1.25} & \colorbox{best2}{1.30} & \colorbox{best2}{1.51} & \colorbox{best2}{1.72} & \colorbox{best2}{2.35} \\
\textbf{FoundDSR}                      &   & \colorbox{best}{0.61} & \colorbox{best}{0.61} & \colorbox{best}{0.63} & \colorbox{best}{0.72} & \colorbox{best}{0.73} & \colorbox{best}{0.80} & \colorbox{best}{0.87} & \colorbox{best}{0.96} & \colorbox{best}{1.07} & \colorbox{best}{1.18} & \colorbox{best}{1.23} & \colorbox{best}{1.43} & \colorbox{best}{1.61} & \colorbox{best}{2.18} \\

\specialrule{1pt}{0.4ex}{0.4ex}
PromptDA       &\multirow{11}{*}{RGB-D-D}  & 2.75 & 2.72 & 2.70 & 2.71 & 2.72 & 2.75 & 2.80 & 2.86 & 2.92 & 2.97 & 2.99 & 3.02 & 3.05 & 3.10 \\
Marigold-DC   &                            & 1.26 & 1.28 & 1.31 & 1.38 & 1.42 & 1.49 & 1.57 & 1.68 & 1.84 & 2.01 & 2.05 & 2.27 & 2.41 & 2.77 \\
InfiniDepth   &                            & 1.08 & 1.09 & 1.11 & 1.16 & 1.21 & 1.30 & 1.35 & 1.46 & 1.65 & 1.78 & 1.90 & 2.02 & 2.18 & 2.72 \\
LDCM          &                            & 0.62 & 0.63 & 0.80 & 0.94 & 0.98 & 1.13 & 1.32 & 1.49 & 1.65 & 1.87 & 1.93 & 2.19 & 2.39 & 2.75 \\
FDSR          &                            & 0.49 & 0.51 & 0.58 & 0.60 & 0.62 & 0.66 & 0.75 & 0.84 & 0.94 & 1.10 & 1.18 & 1.56 & 1.83 & 2.39 \\
DCTNet        &                            & 0.65 & 0.66 & 0.69 & 0.77 & 0.79 & 0.86 & 0.94 & 1.06 & 1.24 & 1.43 & 1.50 & 1.77 & 1.94 & 2.41 \\
DORNet        &                            & 0.49 & 0.50 & 0.53 & 0.62 & 0.64 & 0.72 & 0.79 & 0.89 & 1.03 & 1.18 & 1.23 & 1.47 & 1.67 & 2.24 \\
C2PD          &                            & 0.49 & 0.52 & 0.56 & 0.57 & 0.62 & 0.67 & 0.74 & 0.82 & 0.95 & 1.04 & 1.21 & 1.35 & 1.76 & 2.46 \\
DuCos         &                            & 0.40 & 0.39 & 0.43 & 0.53 & 0.55 & 0.62 & 0.68 & 0.77 & 0.90 & 1.02 & 1.07 & 1.31 & 1.52 & 2.60 \\
\textbf{FoundDSR-S}                    &   & \colorbox{best2}{0.39} & \colorbox{best2}{0.38} & \colorbox{best2}{0.40} & \colorbox{best2}{0.47} & \colorbox{best2}{0.48} & \colorbox{best2}{0.54} & \colorbox{best2}{0.59} & \colorbox{best2}{0.67} & \colorbox{best2}{0.76} & \colorbox{best2}{0.85} & \colorbox{best2}{0.88} & \colorbox{best2}{1.03} & \colorbox{best2}{1.15} & \colorbox{best2}{1.54} \\
\textbf{FoundDSR}                      &   & \colorbox{best}{0.36} & \colorbox{best}{0.36} & \colorbox{best}{0.39} & \colorbox{best}{0.46} & \colorbox{best}{0.47} & \colorbox{best}{0.53} & \colorbox{best}{0.57} & \colorbox{best}{0.64} & \colorbox{best}{0.72} & \colorbox{best}{0.80} & \colorbox{best}{0.83} & \colorbox{best}{0.94} & \colorbox{best}{1.04} & \colorbox{best}{1.34} \\

\specialrule{1pt}{0.4ex}{0.4ex}
PromptDA       &\multirow{10}{*}{iBims-1}  & 13.84 & 14.10 & 14.14 & 14.15 & 14.42 & 13.78 & 14.50 & 14.91 & 15.45 & 15.66 & 15.96 & 16.05 & 16.37 & 17.17 \\
Marigold-DC   &                            & 14.05 & 14.26 & 13.37 & 13.30 & 13.29 & 13.37 & 13.46 & 12.93 & 14.04 & 14.59 & 14.60 & 14.64 & 15.72 & 16.86 \\
LDCM          &                            & 4.78 & 4.62 & 5.11 & 6.45 & 6.79 & 7.40 & 8.01 & 9.05 & 9.81 & 10.64 & 11.05 & 11.96 & 12.36 & 14.17 \\
FDSR          &                            & 2.91 & 3.05 & 3.21 & 3.99 & 4.03 & 4.51 & 4.81 & 5.48 & 6.36 & 7.44 & 8.01 & 9.70 & 10.78 & 13.14 \\
DCTNet        &                            & 3.26 & 3.33 & 3.74 & 4.68 & 4.84 & 5.46 & 5.87 & 6.65 & 7.68 & 8.76 & 9.17 & 10.44 & 11.19 & 13.34 \\
DORNet        &                            & 2.10 & 2.21 & 2.65 & 3.66 & 3.81 & 4.44 & 4.84 & 5.60 & 6.55 & 7.46 & 7.90 & 9.17 & 10.04 & 12.54 \\
C2PD          &                            & 2.13 & 2.38 & 2.50 & 3.16 & 3.43 & 3.95 & 4.22 & 4.91 & 5.95 & 6.60 & 7.51 & 8.32 & 10.21 & 13.25 \\
DuCos         &                            & 2.18 & \colorbox{best2}{2.17} & 2.35 & \colorbox{best2}{3.08} & 3.22 & 3.80 & 4.18 & 4.92 & 5.74 & 6.57 & 7.02 & 8.44 & 9.45 & 12.28 \\
\textbf{FoundDSR-S}                    &   & \colorbox{best2}{2.10} & \colorbox{best}{2.10} & \colorbox{best}{2.24} & \colorbox{best}{2.77} & \colorbox{best}{2.87} & \colorbox{best2}{3.27} & \colorbox{best2}{3.58} & \colorbox{best2}{4.12} & \colorbox{best2}{4.84} & \colorbox{best2}{5.55} & \colorbox{best2}{5.88} & \colorbox{best2}{6.71} & \colorbox{best2}{7.32} & \colorbox{best2}{9.41} \\
\textbf{FoundDSR}                      &   & \colorbox{best}{2.07} & \colorbox{best}{2.10} & \colorbox{best2}{2.27} & \colorbox{best}{2.77} & \colorbox{best2}{2.88} & \colorbox{best}{3.22} & \colorbox{best}{3.47} & \colorbox{best}{3.96} & \colorbox{best}{4.57} & \colorbox{best}{5.19} & \colorbox{best}{5.47} & \colorbox{best}{6.16} & \colorbox{best}{6.67} & \colorbox{best}{8.38} \\

\specialrule{1.5pt}{0.8ex}{0.8ex}
\end{tabular}}
\vspace{-12pt}
\end{table}

\subsection{Comparisons with More Evaluation Metrics}
Tab.~\ref{tab:Supp_MAE} and Tab.~\ref{tab:Supp_d105} further present zero-shot comparisons in terms of MAE and $\delta_{1.05}$ metrics, respectively. The results demonstrate that our FoundDSR consistently exhibits stronger generalization capability than existing DSR methods and foundation models, particularly at large upsampling factors. Compared with the more challenging large-scale settings, smaller upsampling factors require less model capacity to achieve satisfactory performance. Consequently, our lightweight variant, FoundDSR-S, slightly outperforms FoundDSR in some low-scale settings. As the upsampling factor increases, the advantage of the larger FoundDSR becomes increasingly pronounced. For example, at $\times16.0$, FoundDSR reduces the MAE of the second-best method by $26.54\%$, $31.16\%$, $25.81\%$, $31.58\%$, and $29.42\%$ on NYU-v2, Lu, Middlebury, RGB-D-D, and iBims-1, respectively, while improving $\delta_{1.05}$ by $2.23$, $2.84$, $3.95$, $1.61$, and $3.26$ percentage points.

\begin{figure}[!ht]
\centering
\includegraphics[width=1\columnwidth]{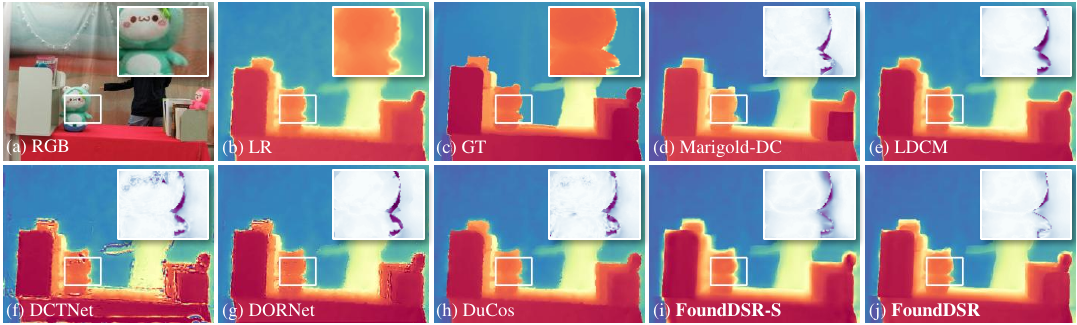}\\
\vspace{-6pt}
\caption{Visual comparisons on the real-world RGB-D-D dataset. The zoomed‑in patches in (d)–(j) highlight the error maps within the white boxes, where darker purple shades indicate higher errors.}\label{fig:real}
\vspace{-8pt}
\end{figure}

\begin{table}[t]
\caption{Quantitative comparisons on five synthetic benchmarks using the $\mathbf{\delta _{1.05}\uparrow}$ metric.}\label{tab:Supp_d105}
\vspace{-8pt}
\centering
\Large
\renewcommand\arraystretch{1.05}
\resizebox{\columnwidth}{!}{
\begin{tabular}{l|c|cccccccccccccc}
\specialrule{1.5pt}{0.8ex}{0.8ex} 
Methods &Datasets  &$\times2.5$ &$\times3.0$ &$\times4.0$     &$\times5.7$ &$\times6.0$ &$\times7.1$  
 &$\times8.0$ &$\times9.3$ &$\times11.0$     &$\times12.5$ &$\times13.0$ &$\times14.9$    &$\times16.0$ &$\times19.3$ \\ \midrule

PromptDA       &\multirow{11}{*}{NYU-v2}   & 92.39 & 92.46 & 92.20 & 92.13 & 92.16 & 92.26 & 91.57 & 91.90 & 91.59 & 91.38 & 91.45 & 91.23 & 90.66 & 90.73 \\
Marigold-DC   &                            & 95.91 & 95.87 & 95.69 & 95.38 & 95.33 & 94.97 & 94.56 & 94.21 & 93.47 & 92.87 & 92.58 & 91.85 & 91.12 & 89.73 \\
InfiniDepth   &                            & 94.37 & 94.21 & 94.06 & 93.80 & 93.59 & 93.29 & 92.79 & 92.74 & 91.63 & 91.15 & 90.81 & 90.31 & 89.30 & 87.52 \\
LDCM          &                            & 98.88 & 98.78 & 98.39 & 97.43 & 97.23 & 96.55 & 95.94 & 95.24 & 94.23 & 93.39 & 93.09 & 92.09 & 91.28 & 89.74 \\
FDSR          &                            & 99.00 & 99.10 & 99.00 & 99.05 & 98.98 & 98.88 & 98.62 & 98.46 & 98.15 & 97.56 & 97.13 & 95.04 & 93.45 & 90.99 \\
DCTNet        &                            & 99.19 & 99.16 & 99.10 & 98.83 & 98.76 & 98.51 & 98.25 & 97.84 & 96.84 & 95.66 & 95.28 & 93.81 & 92.62 & 90.66 \\
DORNet        &                            & 99.12 & 99.13 & 99.09 & 99.07 & 99.02 & 99.04 & 98.75 & 98.53 & 98.22 & 97.71 & 97.23 & 95.10 & 94.34 & 91.72 \\
C2PD          &                            & 99.21 & 99.21 & 99.15 & 99.14 & 99.06 & 98.94 & 98.80 & 98.60 & 98.19 & 97.84 & 97.11 & 96.34 & 93.82 & 90.46 \\
DuCos         &                            & 99.26 & 99.33 & 99.30 & 99.13 & 99.09 & 98.93 & 98.80 & 98.54 & 98.19 & 97.81 & 97.61 & 96.27 & 94.95 & 91.87 \\
\textbf{FoundDSR-S}                    &   & \colorbox{best2}{99.58} & \colorbox{best2}{99.60} & \colorbox{best2}{99.57} & \colorbox{best2}{99.40} & \colorbox{best2}{99.36} & \colorbox{best2}{99.22} & \colorbox{best2}{99.10} & \colorbox{best2}{98.93} & \colorbox{best2}{98.64} & \colorbox{best2}{98.26} & \colorbox{best2}{98.05} & \colorbox{best2}{97.25} & \colorbox{best2}{96.40} & \colorbox{best2}{94.43} \\
\textbf{FoundDSR}                      &   & \colorbox{best}{99.69} & \colorbox{best}{99.69} & \colorbox{best}{99.63} & \colorbox{best}{99.46} & \colorbox{best}{99.43} & \colorbox{best}{99.30} & \colorbox{best}{99.21} & \colorbox{best}{99.04} & \colorbox{best}{98.80} & \colorbox{best}{98.54} & \colorbox{best}{98.39} & \colorbox{best}{97.79} & \colorbox{best}{97.18} & \colorbox{best}{95.51} \\

\specialrule{1pt}{0.4ex}{0.4ex}
Marigold-DC   &\multirow{9}{*}{Lu}         & 74.42 & 74.19 & 74.74 & 73.47 & 73.66 & 73.83 & 73.28 & 78.51 & 73.28 & 72.67 & 73.00 & 76.63 & 71.61 & 69.80 \\
LDCM          &                            & 98.07 & 97.98 & 97.36 & 95.73 & 95.22 & 94.54 & 93.93 & 92.46 & 91.50 & 90.35 & 89.76 & 88.20 & 87.99 & 85.34 \\
FDSR          &                            & 98.06 & 98.19 & 97.90 & 97.80 & 97.73 & 97.48 & 96.99 & 96.57 & 95.76 & 94.66 & 94.11 & 91.68 & 90.49 & 86.92 \\
DCTNet        &                            & 97.91 & 97.88 & 97.74 & 97.29 & 97.17 & 96.72 & 96.23 & 95.32 & 93.96 & 92.57 & 92.08 & 90.07 & 89.36 & 86.20 \\
DORNet        &                            & 98.41 & 98.38 & 98.31 & 97.93 & 97.80 & 97.35 & 96.94 & 96.20 & 95.25 & 94.37 & 94.01 & 92.49 & 91.55 & 87.72 \\
C2PD          &                            & 97.94 & 98.11 & 97.86 & 97.97 & 97.72 & 97.52 & 97.11 & 96.65 & 95.67 & 95.05 & 94.43 & 93.45 & 91.17 & 86.04 \\
DuCos         &                            & 98.67 & 98.73 & 98.63 & 98.32 & 98.19 & 97.93 & 97.66 & 97.10 & 96.30 & 95.64 & 95.26 & 93.30 & 92.18 & 87.85 \\
\textbf{FoundDSR-S}                    &   & \colorbox{best2}{99.13} & \colorbox{best2}{99.18} & \colorbox{best2}{99.17} & \colorbox{best2}{98.91} & \colorbox{best2}{98.84} & \colorbox{best2}{98.58} & \colorbox{best2}{98.36} & \colorbox{best2}{97.91} & \colorbox{best2}{97.33} & \colorbox{best2}{96.56} & \colorbox{best2}{96.21} & \colorbox{best2}{94.86} & \colorbox{best2}{94.24} & \colorbox{best2}{91.16} \\
\textbf{FoundDSR}                      &   & \colorbox{best}{99.34} & \colorbox{best}{99.37} & \colorbox{best}{99.32} & \colorbox{best}{99.02} & \colorbox{best}{98.93} & \colorbox{best}{98.68} & \colorbox{best}{98.44} & \colorbox{best}{98.09} & \colorbox{best}{97.49} & \colorbox{best}{96.92} & \colorbox{best}{96.59} & \colorbox{best}{95.49} & \colorbox{best}{95.02} & \colorbox{best}{92.18} \\

\specialrule{1pt}{0.4ex}{0.4ex}
Marigold-DC   &\multirow{9}{*}{Middlebury} & 90.88 & 90.51 & 89.80 & 88.50 & 88.89 & 87.87 & 87.58 & 86.12 & 85.19 & 83.68 & 83.50 & 81.62 & 81.05 & 77.96 \\
LDCM          &                            & 97.62 & 97.49 & 96.56 & 94.11 & 93.73 & 92.28 & 91.11 & 89.55 & 87.65 & 85.90 & 85.52 & 83.30 & 82.59 & 79.29 \\
FDSR          &                            & 97.77 & 97.71 & 97.33 & 96.96 & 96.83 & 96.49 & 96.03 & 95.50 & 94.41 & 92.51 & 91.68 & 87.54 & 85.55 & 80.72 \\
DCTNet        &                            & 97.28 & 97.25 & 97.08 & 96.44 & 96.31 & 95.57 & 94.86 & 93.49 & 91.37 & 88.97 & 88.26 & 85.28 & 84.04 & 79.95 \\
DORNet        &                            & 97.56 & 97.56 & 97.50 & 97.20 & 97.17 & 96.82 & 96.34 & 95.72 & 94.61 & 93.22 & 92.66 & 89.70 & 88.07 & 82.47 \\
C2PD          &                            & 97.47 & 97.31 & 97.20 & 97.16 & 97.01 & 96.79 & 96.46 & 96.05 & 95.11 & 94.47 & 93.41 & 91.17 & 87.52 & 80.28 \\
DuCos         &                            & 97.86 & 98.00 & 97.90 & 97.53 & 97.45 & 97.13 & 96.78 & 96.29 & 95.66 & 94.90 & 94.35 & 91.50 & 89.58 & 83.42 \\
\textbf{FoundDSR-S}                    &   & \colorbox{best2}{98.53} & \colorbox{best2}{98.56} & \colorbox{best2}{98.46} & \colorbox{best2}{98.10} & \colorbox{best}{98.80} & \colorbox{best2}{97.71} & \colorbox{best2}{97.38} & \colorbox{best2}{96.91} & \colorbox{best2}{96.35} & \colorbox{best2}{95.63} & \colorbox{best2}{95.28} & \colorbox{best2}{93.75} & \colorbox{best2}{92.60} & \colorbox{best2}{88.47} \\
\textbf{FoundDSR}                      &   & \colorbox{best}{98.98} & \colorbox{best}{99.01} & \colorbox{best}{98.86} & \colorbox{best}{98.47} & \colorbox{best2}{98.43} & \colorbox{best}{98.03} & \colorbox{best}{97.72} & \colorbox{best}{97.29} & \colorbox{best}{96.74} & \colorbox{best}{96.12} & \colorbox{best}{95.80} & \colorbox{best}{94.52} & \colorbox{best}{93.53} & \colorbox{best}{89.85} \\

\specialrule{1pt}{0.4ex}{0.4ex}
PromptDA       &\multirow{11}{*}{RGB-D-D}  & 97.13 & 97.11 & 97.02 & 96.78 & 96.76 & 96.59 & 96.46 & 96.31 & 96.32 & 96.05 & 96.11 & 96.05 & 95.91 & 95.75 \\
Marigold-DC   &                            & 98.05 & 97.99 & 97.87 & 97.57 & 97.46 & 97.14 & 96.83 & 96.39 & 95.80 & 95.14 & 94.87 & 94.08 & 93.48 & 92.06 \\
InfiniDepth   &                            & 97.64 & 97.60 & 97.49 & 97.31 & 97.12 & 96.77 & 96.57 & 96.12 & 95.48 & 94.94 & 94.54 & 94.03 & 93.35 & 91.29 \\
LDCM          &                            & 99.20 & 99.17 & 98.70 & 98.05 & 97.90 & 97.34 & 96.76 & 96.14 & 95.45 & 94.66 & 94.45 & 93.51 & 92.86 & 91.42 \\
FDSR          &                            & 99.31 & 99.22 & 99.02 & 99.03 & 98.97 & 98.89 & 98.62 & 98.43 & 98.22 & 97.81 & 97.61 & 96.35 & 95.33 & 93.07 \\
DCTNet        &                            & 99.13 & 99.10 & 98.99 & 98.70 & 98.64 & 98.43 & 98.27 & 98.03 & 97.58 & 97.01 & 96.78 & 95.77 & 95.08 & 93.17 \\
DORNet        &                            & 99.15 & 99.14 & 99.10 & 98.91 & 98.87 & 98.68 & 98.53 & 98.29 & 97.95 & 97.60 & 97.47 & 96.73 & 96.05 & 93.87 \\
C2PD          &                            & 99.28 & 99.14 & 98.95 & 99.05 & 98.89 & 98.82 & 98.55 & 98.39 & 98.14 & 97.90 & 97.53 & 97.09 & 95.75 & 92.97 \\
DuCos         &                            & 99.40 & 99.43 & 99.36 & 99.17 & 99.13 & 98.97 & 98.83 & 98.60 & 98.26 & 97.99 & 97.89 & 97.25 & 96.52 & 93.99 \\
\textbf{FoundDSR-S}                    &   & \colorbox{best2}{99.62} & \colorbox{best2}{99.62} & \colorbox{best2}{99.59} & \colorbox{best2}{99.44} & \colorbox{best2}{99.40} & \colorbox{best2}{99.27} & \colorbox{best2}{99.15} & \colorbox{best2}{98.99} & \colorbox{best2}{98.77} & \colorbox{best2}{98.54} & \colorbox{best2}{98.49} & \colorbox{best2}{98.06} & \colorbox{best2}{97.70} & \colorbox{best2}{96.37} \\
\textbf{FoundDSR}                      &   & \colorbox{best}{99.72} & \colorbox{best}{99.72} & \colorbox{best}{99.66} & \colorbox{best}{99.50} & \colorbox{best}{99.47} & \colorbox{best}{99.34} & \colorbox{best}{99.23} & \colorbox{best}{99.07} & \colorbox{best}{98.87} & \colorbox{best}{98.68} & \colorbox{best}{98.65} & \colorbox{best}{98.35} & \colorbox{best}{98.13} & \colorbox{best}{97.23} \\

\specialrule{1pt}{0.4ex}{0.4ex}
PromptDA       &\multirow{10}{*}{iBims-1}  & 86.92 & 86.86 & 87.07 & 87.33 & 87.08 & 88.26 & 87.57 & 87.65 & 87.18 & 87.25 & 87.22 & 86.99 & 86.72 & 86.38 \\
Marigold-DC   &                            & 86.50 & 86.51 & 86.37 & 85.83 & 85.74 & 85.64 & 84.98 & 85.85 & 84.27 & 83.47 & 83.61 & 83.49 & 81.79 & 80.76 \\
LDCM          &                            & 93.68 & 93.70 & 92.89 & 91.32 & 91.09 & 90.19 & 89.56 & 88.60 & 87.51 & 86.52 & 86.19 & 85.13 & 84.50 & 82.60 \\
FDSR          &                            & 95.99 & 95.84 & 95.64 & 94.92 & 94.93 & 94.50 & 94.27 & 93.63 & 92.90 & 91.87 & 91.24 & 89.02 & 87.63 & 84.71 \\
DCTNet        &                            & 94.55 & 94.54 & 94.24 & 93.48 & 93.35 & 92.81 & 92.38 & 91.71 & 90.41 & 88.90 & 88.41 & 86.70 & 85.71 & 83.30 \\
DORNet        &                            & 96.34 & 96.22 & 95.83 & 95.15 & 95.03 & 94.55 & 94.15 & 93.63 & 92.78 & 92.01 & 91.63 & 90.05 & 88.93 & 85.62 \\
C2PD          &                            & 96.39 & 96.35 & 96.47 & 95.80 & 95.60 & 95.10 & 94.93 & 94.35 & 93.55 & 93.00 & 92.05 & 91.22 & 88.84 & 84.81 \\
DuCos         &                            & 96.70 & 96.49 & \colorbox{best2}{97.16} & 96.37 & 96.12 & 95.69 & 95.30 & 94.66 & 93.98 & 93.30 & 92.94 & 91.11 & 89.79 & 86.15 \\
\textbf{FoundDSR-S}                    &   & \colorbox{best2}{96.80} & \colorbox{best2}{96.81} & \colorbox{best}{96.78} & \colorbox{best}{96.48} & \colorbox{best}{96.43} & \colorbox{best}{96.09} & \colorbox{best}{95.82} & \colorbox{best}{95.40} & \colorbox{best}{94.83} & \colorbox{best2}{94.23} & \colorbox{best2}{93.94} & \colorbox{best2}{93.03} & \colorbox{best2}{92.39} & \colorbox{best2}{90.01} \\
\textbf{FoundDSR}                      &   & \colorbox{best}{97.12} & \colorbox{best}{97.05} & \colorbox{best}{96.78} & \colorbox{best2}{96.29} & \colorbox{best2}{96.22} & \colorbox{best2}{95.79} & \colorbox{best2}{95.61} & \colorbox{best2}{95.20} & \colorbox{best2}{94.73} & \colorbox{best}{94.28} & \colorbox{best}{94.15} & \colorbox{best}{93.55} & \colorbox{best}{93.05} & \colorbox{best}{91.40} \\

\specialrule{1.5pt}{0.8ex}{0.8ex}
\end{tabular}}
\vspace{-12pt}
\end{table}

\subsection{More Visual Comparisons}

Fig.~\ref{fig:real} shows qualitative comparisons on real-world scenarios with unseen degradations. The results further confirm the strong generalization capability of our method in reconstructing high-quality depth under challenging real-world conditions. For example, the toy restored by FoundDSR exhibits more accurate depth structures than those produced by competing methods.

\begin{figure}[!ht]
\centering
\includegraphics[width=1\columnwidth]{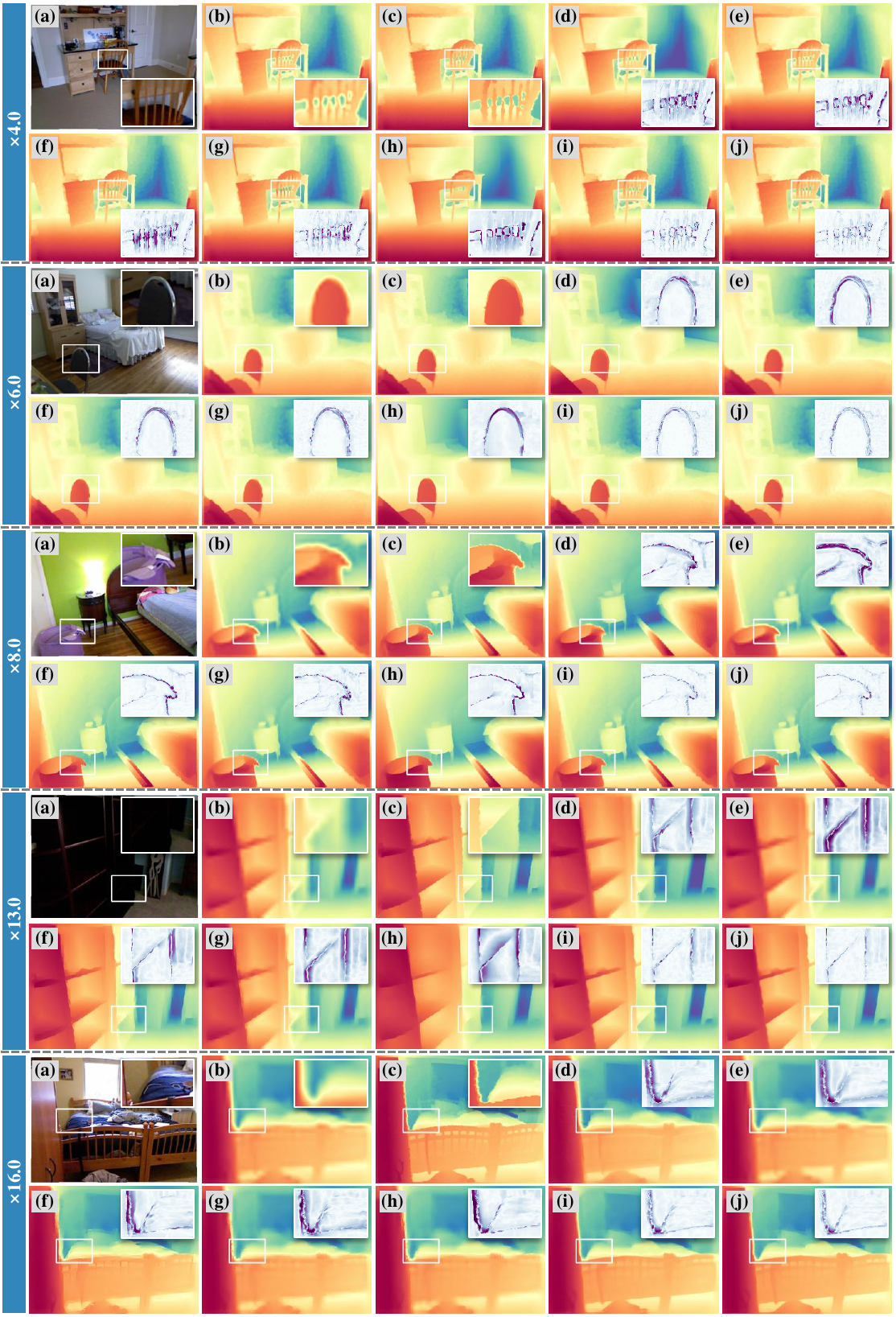}\\
\vspace{-3pt}
\caption{Visual comparisons on NYU-v2 across integer scales. (a)–(j) denote RGB, LR, GT, Marigold-DC, LDCM, DCTNet, DORNet, DuCos, FoundDSR-S, and FoundDSR, respectively.}\label{fig:sup_integer}
\vspace{-15pt}
\end{figure}

\begin{figure}[!ht]
\centering
\includegraphics[width=1\columnwidth]{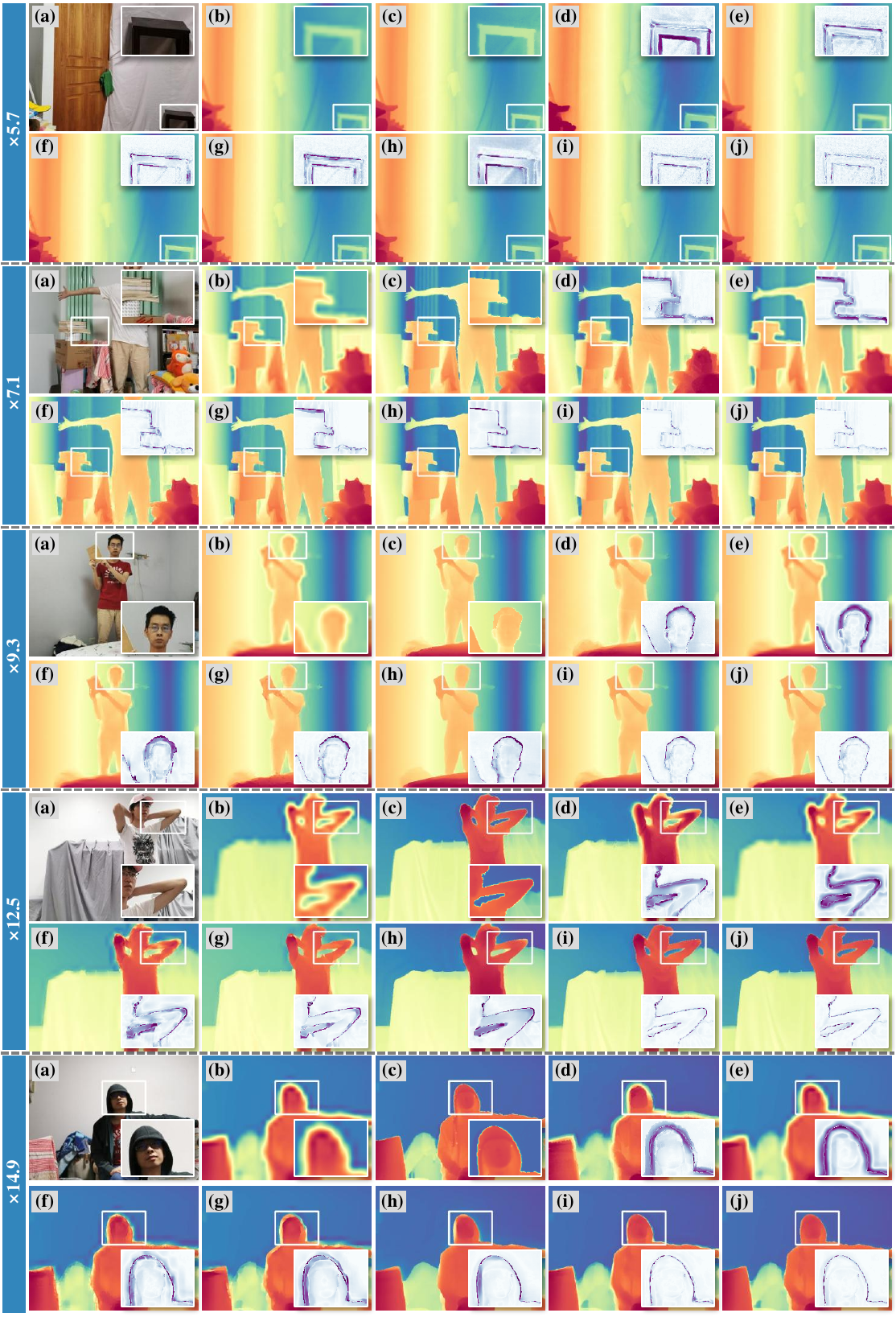}\\
\vspace{-8pt}
\caption{Visual comparisons on RGB-D-D across non-integer scales. (a)–(j) denote RGB, LR, GT, Marigold-DC, LDCM, DCTNet, DORNet, DuCos, FoundDSR-S, and FoundDSR, respectively.}\label{fig:sup_noninteger}
\vspace{-20pt}
\end{figure}

Figs.~\ref{fig:sup_integer} and \ref{fig:sup_noninteger} present qualitative comparisons and corresponding error maps under integer and non-integer upsampling factors, respectively. FoundDSR consistently delivers superior reconstruction quality over existing DSR methods and depth foundation models across different scales, recovering more accurate depth structures with lower errors. For example, the cabinet at $\times13.0$ in Fig.~\ref{fig:sup_integer} and the arm at $\times12.5$ in Fig.~\ref{fig:sup_noninteger} reconstructed by our method exhibit more faithful geometric structures and are closer to the GT depth than those of competing methods. 

\begin{figure}[t]
\centering
\includegraphics[width=1\columnwidth]{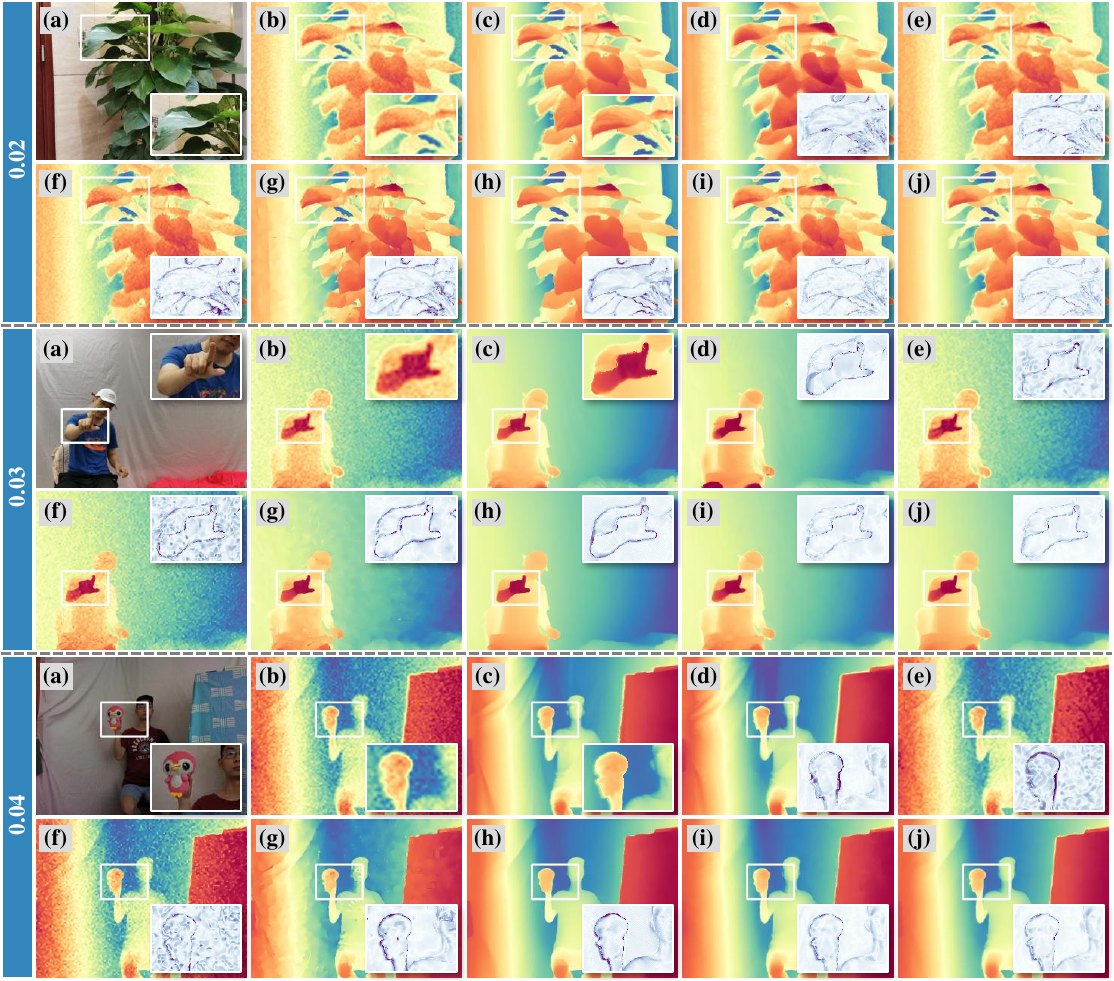}\\
\vspace{-3pt}
\caption{Visual comparisons on $\times4.0$ RGB-D-D under various noise levels. (a)–(j): RGB, LR, GT, Marigold-DC, LDCM, DCTNet, DORNet, DuCos, FoundDSR-S, and FoundDSR.}\label{fig:sup_noise}
\vspace{-6pt}
\end{figure}

Furthermore, Fig.~\ref{fig:sup_noise} provides visual comparisons under different levels of zero-mean Gaussian noise, with standard deviations ranging from $0.02$ to $0.04$. FoundDSR remains robust under increasing noise levels, effectively suppressing noise while recovering accurate geometric details. For instance, the arm reconstructed by our method in Fig.~\ref{fig:sup_noise} contains less residual noise and lower reconstruction errors than other approaches. These results further demonstrate the robustness and strong zero-shot generalization of our FoundDSR.

\subsection{More Ablation Study}

\begin{wrapfigure}{r}{0.5\textwidth}
\centering
\includegraphics[width=\linewidth]{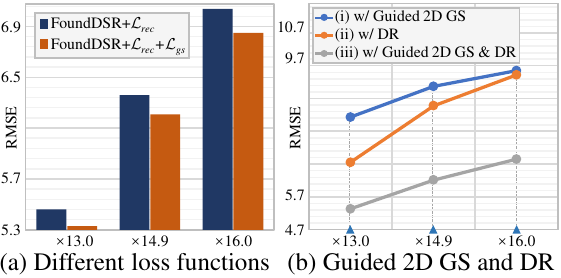}
\caption{Ablation study of (a) different losses and (b) guided 2D GS and depth reconstruction (DR) on $\times13.0$, $\times14.9$, and $\times16.0$ NYU-v2.}
\label{fig:ab_loss_branch}
\end{wrapfigure}

\noindent \textbf{Different Loss Functions.} 
Fig.~\ref{fig:ab_loss_branch}(a) presents the ablation study on different loss functions, including the standard reconstruction loss $\mathcal{L}_{rec}$ and our Gaussian loss $\mathcal{L}_{gs}$. While $\mathcal{L}_{rec}$ alone already yields 
a strong baseline, incorporating $\mathcal{L}_{gs}$ consistently reduces the RMSE across all upsampling factors, with more pronounced improvements at higher scales. Specifically, adding $\mathcal{L}_{gs}$ decreases RMSE by $0.13cm$, $0.15cm$, and $0.19cm$ at upsampling scales of $\times13.0$, $\times14.9$, and $\times16.0$, respectively.  These results demonstrate that $\mathcal{L}_{gs}$ effectively improves the Gaussian-upsampled representation and facilitates more accurate depth reconstruction.

\noindent \textbf{Guided 2D GS and Depth Reconstruction Branches.} 
Fig.~\ref{fig:ab_loss_branch}(b) analyzes the contributions of the guided 2D GS and depth reconstruction (DR) branches. The results show that guided 2D GS (i) stabilizes performance  across different scale factors, reducing the performance gap between continuous scales. In contrast, the DR branch (ii) substantially improves reconstruction accuracy at every scale, but exhibits larger fluctuations across continuous scale factors. When both branches are combined, FoundDSR (iii) achieves the best performance across all evaluated scales, reducing RMSE by $2.80cm$ and $1.42cm$ compared with (i) and (ii), respectively, on $\times16.0$ NYU-v2.






\end{document}

%% file: math_commands.tex
\usepackage{amsmath,amsfonts,bm}

\def\eqref#1{equation~\ref{#1}}

\def\1{\bm{1}}

\DeclareMathAlphabet{\mathsfit}{\encodingdefault}{\sfdefault}{m}{sl}
\SetMathAlphabet{\mathsfit}{bold}{\encodingdefault}{\sfdefault}{bx}{n}



%% file: iclr2027_conference.bib
@inproceedings{wang2023learning,
  title={Learning continuous depth representation via geometric spatial aggregator},
  author={Wang, Xiaohang and Chen, Xuanhong and Ni, Bingbing and Tong, Zhengyan and Wang, Hang},
  booktitle={AAAI},
  pages={2698--2706},
  year={2023}
}

@inproceedings{chen2024intrinsic,
  title={Intrinsic Phase-Preserving Networks for Depth Super Resolution},
  author={Chen, Xuanhong and Wang, Hang and Chen, Jialiang and Feng, Kairui and Liu, Jinfan and Wang, Xiaohang and Zhang, Weimin and Ni, Bingbing},
  booktitle={AAAI},
  pages={1210--1218},
  year={2024}
}

@article{yang2022codon,
  title={CODON: On orchestrating cross-domain attentions for depth super-resolution},
  author={Yang, Yuxiang and Cao, Qi and Zhang, Jing and Tao, Dacheng},
  journal={International Journal of Computer Vision},
  volume={130},
  number={2},
  pages={267--284},
  year={2022},
  publisher={Springer}
}

@article{wang2026scene,
  title={Scene Prior Filtering for Depth Super-Resolution: Zhengxue Wang et al.},
  author={Wang, Zhengxue and Yan, Zhiqiang and Yang, Ming-Hsuan and Pan, Jinshan and Gao, Guangwei and Tai, Ying and Yang, Jian},
  journal={International Journal of Computer Vision},
  volume={134},
  number={5},
  pages={251},
  year={2026},
  publisher={Springer}
}

@inproceedings{zhong2026dual,
  title={Dual Graph Regularized Deep Unfolding Network for Guided Depth Map Super-resolution},
  author={Zhong, Zhiwei and Chen, Peilin and Shen, Qiangqiang and Li, Bo and Wang, Shiqi},
  booktitle={CVPR},
  pages={16322--16332},
  year={2026}
}

@inproceedings{yan2025ducos,
  title={Ducos: Duality constrained depth super-resolution via foundation model},
  author={Yan, Zhiqiang and Wang, Zhengxue and Dong, Haoye and Li, Jun and Yang, Jian and Lee, Gim Hee},
  booktitle={ICCV},
  pages={8361--8371},
  year={2025}
}

@article{zhong2025dual,
  title={Dual-level cross-modality neural architecture search for guided image super-resolution},
  author={Zhong, Zhiwei and Liu, Xianming and Jiang, Junjun and Zhao, Debin and Wang, Shiqi},
  journal={IEEE Transactions on Pattern Analysis and Machine Intelligence},
  year={2025},
  publisher={IEEE}
}

@inproceedings{he2021towards,
  title={Towards fast and accurate real-world depth super-resolution: Benchmark dataset and baseline},
  author={He, Lingzhi and Zhu, Hongguang and Li, Feng and Bai, Huihui and Cong, Runmin and Zhang, Chunjie and Lin, Chunyu and Liu, Meiqin and Zhao, Yao},
  booktitle={CVPR},
  pages={9229--9238},
  year={2021}
}

@inproceedings{wang2025dornet,
  title={DORNet: A Degradation Oriented and Regularized Network for Blind Depth Super-Resolution},
  author={Wang, Zhengxue and Yan, Zhiqiang and Pan, Jinshan and Gao, Guangwei and Zhang, Kai and Yang, Jian},
  booktitle={CVPR},
  pages={15813--15822},
  year={2025}
}

@inproceedings{song2020channel,
  title={Channel attention based iterative residual learning for depth map super-resolution},
  author={Song, Xibin and Dai, Yuchao and Zhou, Dingfu and Liu, Liu and Li, Wei and Li, Hongdong and Yang, Ruigang},
  booktitle={CVPR},
  pages={5631--5640},
  year={2020}
}

@article{kim2021deformable,
  title={Deformable kernel networks for joint image filtering},
  author={Kim, Beomjun and Ponce, Jean and Ham, Bumsub},
  journal={International Journal of Computer Vision},
  volume={129},
  number={2},
  pages={579--600},
  year={2021},
  publisher={Springer}
}

@article{zhong2023deep,
  title={Deep attentional guided image filtering},
  author={Zhong, Zhiwei and Liu, Xianming and Jiang, Junjun and Zhao, Debin and Ji, Xiangyang},
  journal={IEEE Transactions on Neural Networks and Learning Systems},
  year={2023},
  publisher={IEEE}
}

@inproceedings{zheng2025decoupling,
  title={Decoupling fine detail and global geometry for compressed depth map super-resolution},
  author={Zheng, Huan and Han, Wencheng and Shen, Jianbing},
  booktitle={CVPR},
  pages={951--960},
  year={2025}
}

@article{deng2020deep,
  title={Deep convolutional neural network for multi-modal image restoration and fusion},
  author={Deng, Xin and Dragotti, Pier Luigi},
  journal={IEEE Transactions on Pattern Analysis and Machine Intelligence},
  volume={43},
  number={10},
  pages={3333--3348},
  year={2020},
  publisher={IEEE}
}

@inproceedings{zhao2022discrete,
  title={Discrete cosine transform network for guided depth map super-resolution},
  author={Zhao, Zixiang and Zhang, Jiangshe and Xu, Shuang and Lin, Zudi and Pfister, Hanspeter},
  booktitle={CVPR},
  pages={5697--5707},
  year={2022}
}

@inproceedings{zhao2023spherical,
  title={Spherical space feature decomposition for guided depth map super-resolution},
  author={Zhao, Zixiang and Zhang, Jiangshe and Gu, Xiang and Tan, Chengli and Xu, Shuang and Zhang, Yulun and Timofte, Radu and Van Gool, Luc},
  booktitle={ICCV},
  pages={12547--12558},
  year={2023}
}

@inproceedings{sun2021learning,
  title={Learning scene structure guidance via cross-task knowledge transfer for single depth super-resolution},
  author={Sun, Baoli and Ye, Xinchen and Li, Baopu and Li, Haojie and Wang, Zhihui and Xu, Rui},
  booktitle={CVPR},
  pages={7792--7801},
  year={2021}
}

@inproceedings{yuan2023structure,
  title={Structure flow-guided network for real depth super-resolution},
  author={Yuan, Jiayi and Jiang, Haobo and Li, Xiang and Qian, Jianjun and Li, Jun and Yang, Jian},
  booktitle={AAAI},
  pages={3340--3348},
  year={2023}
}

@article{wang2023rgb,
  title={RGB-guided depth map recovery by two-stage coarse-to-fine dense CRF models},
  author={Wang, Haotian and Yang, Meng and Zhu, Ce and Zheng, Nanning},
  journal={IEEE Transactions on Image Processing},
  volume={32},
  pages={1315--1328},
  year={2023},
  publisher={IEEE}
}

@article{yan2026event,
  title={Event-driven dynamic scene depth completion},
  author={Yan, Zhiqiang and Jiao, Jianhao and Wang, Zhengxue and Lee, Gim Hee},
  journal={NeurIPS},
  volume={38},
  pages={142971--142993},
  year={2026}
}

@article{yan2025tri,
  title={Tri-Perspective View Decomposition for Geometry Aware Depth Completion and Super-Resolution},
  author={Yan, Zhiqiang and Wang, Kun and Li, Xiang and Gao, Guangwei and Li, Jun and Yang, Jian},
  journal={IEEE Transactions on Pattern Analysis and Machine Intelligence},
  year={2025},
  publisher={IEEE}
}

@inproceedings{kang2025c2pd,
  title={C2pd: Continuity-constrained pixelwise deformation for guided depth super-resolution},
  author={Kang, Jiahui and Cai, Qing and Tan, Runqing and Liu, Yimei and Liu, Zhi},
  booktitle={AAAI},
  pages={4212--4220},
  year={2025}
}

@inproceedings{wang2024sgnet,
  title={Sgnet: Structure guided network via gradient-frequency awareness for depth map super-resolution},
  author={Wang, Zhengxue and Yan, Zhiqiang and Yang, Jian},
  booktitle={AAAI},
  pages={5823--5831},
  year={2024}
}

@inproceedings{roberts2021hypersim,
  title={Hypersim: A photorealistic synthetic dataset for holistic indoor scene understanding},
  author={Roberts, Mike and Ramapuram, Jason and Ranjan, Anurag and Kumar, Atulit and Bautista, Miguel Angel and Paczan, Nathan and Webb, Russ and Susskind, Joshua M},
  booktitle={ICCV},
  pages={10912--10922},
  year={2021}
}

@inproceedings{ye2025semantics,
  title={Semantics-Driven Contrastive Learning for Real-World Depth Super Resolution},
  author={Ye, Xinchen and Zhang, Aokai and Xu, Rui},
  booktitle={ACMMM},
  pages={3085--3093},
  year={2025}
}

@inproceedings{kirillov2023segment,
  title={Segment anything},
  author={Kirillov, Alexander and Mintun, Eric and Ravi, Nikhila and Mao, Hanzi and Rolland, Chloe and Gustafson, Laura and Xiao, Tete and Whitehead, Spencer and Berg, Alexander C and Lo, Wan-Yen and others},
  booktitle={ICCV},
  pages={4015--4026},
  year={2023}
}

@inproceedings{yang2024depth,
  title={Depth anything: Unleashing the power of large-scale unlabeled data},
  author={Yang, Lihe and Kang, Bingyi and Huang, Zilong and Xu, Xiaogang and Feng, Jiashi and Zhao, Hengshuang},
  booktitle={CVPR},
  pages={10371--10381},
  year={2024}
}

@article{yang2024depthv2,
  title={Depth anything v2},
  author={Yang, Lihe and Kang, Bingyi and Huang, Zilong and Zhao, Zhen and Xu, Xiaogang and Feng, Jiashi and Zhao, Hengshuang},
  journal={NeurIPS},
  volume={37},
  pages={21875--21911},
  year={2024}
}

@inproceedings{ke2024repurposing,
  title={Repurposing diffusion-based image generators for monocular depth estimation},
  author={Ke, Bingxin and Obukhov, Anton and Huang, Shengyu and Metzger, Nando and Daudt, Rodrigo Caye and Schindler, Konrad},
  booktitle={CVPR},
  pages={9492--9502},
  year={2024}
}

@inproceedings{viola2025marigold,
  title={Marigold-dc: Zero-shot monocular depth completion with guided diffusion},
  author={Viola, Massimiliano and Qu, Kevin and Metzger, Nando and Ke, Bingxin and Becker, Alexander and Schindler, Konrad and Obukhov, Anton},
  booktitle={ICCV},
  pages={5359--5370},
  year={2025}
}

@inproceedings{shi2025fedawa,
  title={FedAWA: adaptive optimization of aggregation weights in federated learning using client vectors},
  author={Shi, Changlong and Zhao, He and Zhang, Bingjie and Zhou, Mingyuan and Guo, Dandan and Chang, Yi},
  booktitle={CVPR},
  pages={30651--30660},
  year={2025}
}

@inproceedings{lin2025prompting,
  title={Prompting depth anything for 4k resolution accurate metric depth estimation},
  author={Lin, Haotong and Peng, Sida and Chen, Jingxiao and Peng, Songyou and Sun, Jiaming and Liu, Minghuan and Bao, Hujun and Feng, Jiashi and Zhou, Xiaowei and Kang, Bingyi},
  booktitle={CVPR},
  pages={17070--17080},
  year={2025}
}

@article{loshchilov2017decoupled,
  title={Decoupled weight decay regularization},
  author={Loshchilov, Ilya and Hutter, Frank},
  journal={arXiv preprint arXiv:1711.05101},
  year={2017}
}

@article{yu2026large,
  title={Large Depth Completion Model from Sparse Observations},
  author={Yu, Zhu and Zhao, Zhengyi and Zhang, Runmin and Qiu, Lingteng and Qiu, Kejie and He, Yisheng and Zhu, Siyu and Dong, Zilong and Cao, Si-Yuan and Shen, Hui-Liang},
  journal={arXiv preprint arXiv:2605.30115},
  year={2026}
}

@article{yu2026infinidepth,
  title={InfiniDepth: Arbitrary-Resolution and Fine-Grained Depth Estimation with Neural Implicit Fields},
  author={Yu, Hao and Lin, Haotong and Wang, Jiawei and Li, Jiaxin and Wang, Yida and Zhang, Xueyang and Wang, Yue and Zhou, Xiaowei and Hu, Ruizhen and Peng, Sida},
  journal={arXiv preprint arXiv:2601.03252},
  year={2026}
}

@article{hu2024metric3d,
  title={Metric3d v2: A versatile monocular geometric foundation model for zero-shot metric depth and surface normal estimation},
  author={Hu, Mu and Yin, Wei and Zhang, Chi and Cai, Zhipeng and Long, Xiaoxiao and Chen, Hao and Wang, Kaixuan and Yu, Gang and Shen, Chunhua and Shen, Shaojie},
  journal={IEEE Transactions on Pattern Analysis and Machine Intelligence},
  volume={46},
  number={12},
  pages={10579--10596},
  year={2024},
  publisher={IEEE}
}

@article{kerbl20233d,
  title={3d gaussian splatting for real-time radiance field rendering.},
  author={Kerbl, Bernhard and Kopanas, Georgios and Leimk{\"u}hler, Thomas and Drettakis, George and others},
  journal={ACM Trans. Graph.},
  volume={42},
  number={4},
  pages={139--1},
  year={2023}
}

@inproceedings{xu2025depthsplat,
  title={Depthsplat: Connecting gaussian splatting and depth},
  author={Xu, Haofei and Peng, Songyou and Wang, Fangjinhua and Blum, Hermann and Barath, Daniel and Geiger, Andreas and Pollefeys, Marc},
  booktitle={CVPR},
  pages={16453--16463},
  year={2025}
}

@inproceedings{peng2026pixel,
  title={Pixel to gaussian: Ultra-fast continuous super-resolution with 2d gaussian modeling},
  author={Peng, Long and Wu, Anran and Li, Wenbo and Zhang, Xinjie and Dai, Xueyuan and Di, Xin and Sun, Haoze and Pei, Renjing and Wang, Yang and Cao, Yang and others},
  booktitle={ICLR},
  volume={2026},
  pages={92172--92199},
  year={2026}
}

@inproceedings{ranftl2021vision,
  title={Vision transformers for dense prediction},
  author={Ranftl, Ren{\'e} and Bochkovskiy, Alexey and Koltun, Vladlen},
  booktitle={ICCV},
  pages={12179--12188},
  year={2021}
}

@article{oquab2023dinov2,
  title={Dinov2: Learning robust visual features without supervision},
  author={Oquab, Maxime and Darcet, Timoth{\'e}e and Moutakanni, Th{\'e}o and Vo, Huy and Szafraniec, Marc and Khalidov, Vasil and Fernandez, Pierre and Haziza, Daniel and Massa, Francisco and El-Nouby, Alaaeldin and others},
  journal={arXiv preprint arXiv:2304.07193},
  year={2023}
}

@article{ye2025gsplat,
  title={gsplat: An open-source library for Gaussian splatting},
  author={Ye, Vickie and Li, Ruilong and Kerr, Justin and Turkulainen, Matias and Yi, Brent and Pan, Zhuoyang and Seiskari, Otto and Ye, Jianbo and Hu, Jeffrey and Tancik, Matthew and others},
  journal={Journal of Machine Learning Research},
  volume={26},
  number={34},
  pages={1--17},
  year={2025}
}

@article{wang2026graph,
  title={Graph-Aware Prompting Network for Generalizable Depth Super-Resolution},
  author={Wang, Zhengxue and Liu, Lina and Li, Xiang and Yan, Zhiqiang and Yang, Jian},
  journal={IEEE Transactions on Circuits and Systems for Video Technology},
  year={2026},
  publisher={IEEE}
}

@inproceedings{zhang2024gaussianimage,
  title={Gaussianimage: 1000 fps image representation and compression by 2d gaussian splatting},
  author={Zhang, Xinjie and Ge, Xingtong and Xu, Tongda and He, Dailan and Wang, Yan and Qin, Hongwei and Lu, Guo and Geng, Jing and Zhang, Jun},
  booktitle={ECCV},
  pages={327--345},
  year={2024},
  organization={Springer}
}

@inproceedings{chen2025generalized,
  title={Generalized and efficient 2d gaussian splatting for arbitrary-scale super-resolution},
  author={Chen, Du and Chen, Liyi and Zhang, Zhengqiang and Zhang, Lei},
  booktitle={ICCV},
  pages={26435--26445},
  year={2025}
}

@inproceedings{zeng2025instant,
  title={Instant GaussianImage: A generalizable and self-adaptive image representation via 2D Gaussian splatting},
  author={Zeng, Zhaojie and Wang, Yuesong and Guan, Tao and Yang, Chao and Ju, Lili},
  booktitle={ICCV},
  pages={27896--27905},
  year={2025}
}

@article{cho2021deep,
  title={Deep monocular depth estimation leveraging a large-scale outdoor stereo dataset},
  author={Cho, Jaehoon and Min, Dongbo and Kim, Youngjung and Sohn, Kwanghoon},
  journal={Expert Systems with Applications},
  volume={178},
  pages={114877},
  year={2021},
  publisher={Elsevier}
}

@article{kim2016structure,
  title={Structure selective depth superresolution for RGB-D cameras},
  author={Kim, Youngjung and Ham, Bumsub and Oh, Changjae and Sohn, Kwanghoon},
  journal={IEEE Transactions on Image Processing},
  volume={25},
  number={11},
  pages={5227--5238},
  year={2016},
  publisher={IEEE}
}

@inproceedings{kim2017deep,
  title={Deep stereo confidence prediction for depth estimation},
  author={Kim, Sunok and Min, Dongbo and Ham, Bumsub and Kim, Seungryong and Sohn, Kwanghoon},
  booktitle={ICIP},
  pages={992--996},
  year={2017}
}

@article{kim2018deep,
  title={Deep monocular depth estimation via integration of global and local predictions},
  author={Kim, Youngjung and Jung, Hyungjoo and Min, Dongbo and Sohn, Kwanghoon},
  journal={IEEE Transactions on Image Processing},
  volume={27},
  number={8},
  pages={4131--4144},
  year={2018},
  publisher={IEEE}
}

@inproceedings{wang2026spatiotemporal,
  title={Spatiotemporal difference network for video depth super-resolution},
  author={Wang, Zhengxue and Wu, Yuan and Li, Xiang and Yan, Zhiqiang and Yang, Jian},
  booktitle={AAAI},
  pages={10403--10411},
  year={2026}
}

@article{tan2026masked,
  title={Masked depth modeling for spatial perception},
  author={Tan, Bin and Sun, Changjiang and Qin, Xiage and Adai, Hanat and Fu, Zelin and Zhou, Tianxiang and Zhang, Han and Xu, Yinghao and Zhu, Xing and Shen, Yujun and others},
  journal={arXiv preprint arXiv:2601.17895},
  year={2026}
}

@inproceedings{karaev2023dynamicstereo,
  title={Dynamicstereo: Consistent dynamic depth from stereo videos},
  author={Karaev, Nikita and Rocco, Ignacio and Graham, Benjamin and Neverova, Natalia and Vedaldi, Andrea and Rupprecht, Christian},
  booktitle={CVPR},
  pages={13229--13239},
  year={2023}
}

@article{replica19arxiv,
  title =   {The {R}eplica Dataset: A Digital Replica of Indoor Spaces},
  author =  {Julian Straub and Thomas Whelan and Lingni Ma and Yufan Chen and Erik Wijmans and Simon Green and Jakob J. Engel and Raul Mur-Artal and Carl Ren and Shobhit Verma and Anton Clarkson and Mingfei Yan and Brian Budge and Yajie Yan and Xiaqing Pan and June Yon and Yuyang Zou and Kimberly Leon and Nigel Carter and Jesus Briales and  Tyler Gillingham and  Elias Mueggler and Luis Pesqueira and Manolis Savva and Dhruv Batra and Hauke M. Strasdat and Renzo De Nardi and Michael Goesele and Steven Lovegrove and Richard Newcombe },
  journal = {arXiv preprint arXiv:1906.05797},
  year =    {2019}
}

@inproceedings{koch2018evaluation,
  title={Evaluation of cnn-based single-image depth estimation methods},
  author={Koch, Tobias and Liebel, Lukas and Fraundorfer, Friedrich and K{\"o}rner, Marco},
  booktitle={ECCV},
  pages={331--348},
  year={2018}
}

@inproceedings{silberman2012indoor,
  title={Indoor segmentation and support inference from rgbd images},
  author={Silberman, Nathan and Hoiem, Derek and Kohli, Pushmeet and Fergus, Rob},
  booktitle={ECCV},
  pages={746--760},
  year={2012},
  organization={Springer}
}

@inproceedings{lu2014depth,
  title={Depth enhancement via low-rank matrix completion},
  author={Lu, Si and Ren, Xiaofeng and Liu, Feng},
  booktitle={CVPR},
  pages={3390--3397},
  year={2014}
}

@inproceedings{hirschmuller2007evaluation,
  title={Evaluation of cost functions for stereo matching},
  author={Hirschmuller, Heiko and Scharstein, Daniel},
  booktitle={CVPR},
  pages={1--8},
  year={2007},
  organization={IEEE}
}

@inproceedings{scharstein2007learning,
  title={Learning conditional random fields for stereo},
  author={Scharstein, Daniel and Pal, Chris},
  booktitle={CVPR},
  pages={1--8},
  year={2007},
}
